\pdfoutput=1

\documentclass[11pt]{article}

\usepackage[preprint]{acl}

\usepackage{times}
\usepackage{latexsym}

\usepackage[T1]{fontenc}

\usepackage[utf8]{inputenc}

\usepackage{microtype}

\usepackage{inconsolata}

\usepackage{wrapfig}
\usepackage{graphicx}
\usepackage{booktabs}
\usepackage{amsmath}
\usepackage{amssymb}
\usepackage{microtype}
\usepackage{pifont}
\usepackage{colortbl}
\usepackage{color}
\usepackage{placeins}
\usepackage{booktabs} 
\usepackage{float}
\usepackage{soul}
\usepackage{enumitem}
\usepackage{adjustbox}
\usepackage{commath}
\usepackage{bbm}
\usepackage{multirow}
\usepackage{caption}
\usepackage{longtable}
\usepackage{array}
\usepackage{xcolor}
\usepackage{makecell}
\usepackage{subcaption}
\usepackage{tcolorbox}
\usepackage{diagbox}

\usepackage{algorithm}
\usepackage{algorithmic}

\title{Scaling Text-Rich Image Understanding \\ via Code-Guided Synthetic Multimodal Data Generation}

\author{Yue Yang$^{*1}$, Ajay Patel$^{*1}$, Matt Deitke$^{2}$, Tanmay Gupta$^{2}$, Luca Weihs$^{2}$, \\ \textbf{Andrew Head$^{1}$,  Mark Yatskar$^{1}$, Chris Callison-Burch$^{1}$, } \\ \textbf{Ranjay Krishna$^{2}$, Aniruddha Kembhavi$^{2}$, Christopher Clark$^{2}$}\\
$^{1}$University of Pennsylvania, $^{2}$Allen Institute for Artificial Intelligence\\
 {\small $^*$ Equal Contribution \hspace{.5cm} \tt{\{yueyang1, ajayp\}@seas.upenn.edu} \hspace{.5cm} \href{https://yueyang1996.github.io/cosyn/}{yueyang1996.github.io/cosyn} } }

\begin{document}
\maketitle
\begin{abstract}
Reasoning about images with rich text, such as charts and documents, is a critical application of vision-language models (VLMs). 
However, VLMs often struggle in these domains due to the scarcity of diverse text-rich vision-language data. 
To address this challenge, we present CoSyn, a framework that leverages the coding capabilities of text-only large language models (LLMs) to automatically create synthetic text-rich multimodal data. 
Given input text describing a target domain (e.g., ``nutrition fact labels''), CoSyn prompts an LLM to generate code (Python, HTML, LaTeX, etc.) for rendering synthetic images. 
With the underlying code as textual representations of the synthetic images, CoSyn can generate high-quality instruction-tuning data, again relying on a text-only LLM.
Using CoSyn, we constructed a dataset comprising 400K images and 2.7M rows of vision-language instruction-tuning data. 
Comprehensive experiments on seven benchmarks demonstrate that models trained on our synthetic data achieve state-of-the-art performance among competitive open-source models, including Llama 3.2, and surpass proprietary models such as GPT-4V and Gemini 1.5 Flash. 
Furthermore, CoSyn can produce synthetic pointing data, enabling VLMs to ground information within input images, showcasing its potential for developing multimodal agents capable of acting in real-world environments.

\end{abstract}

\begin{figure}[!t]
    \centering
    \includegraphics[width=\columnwidth]{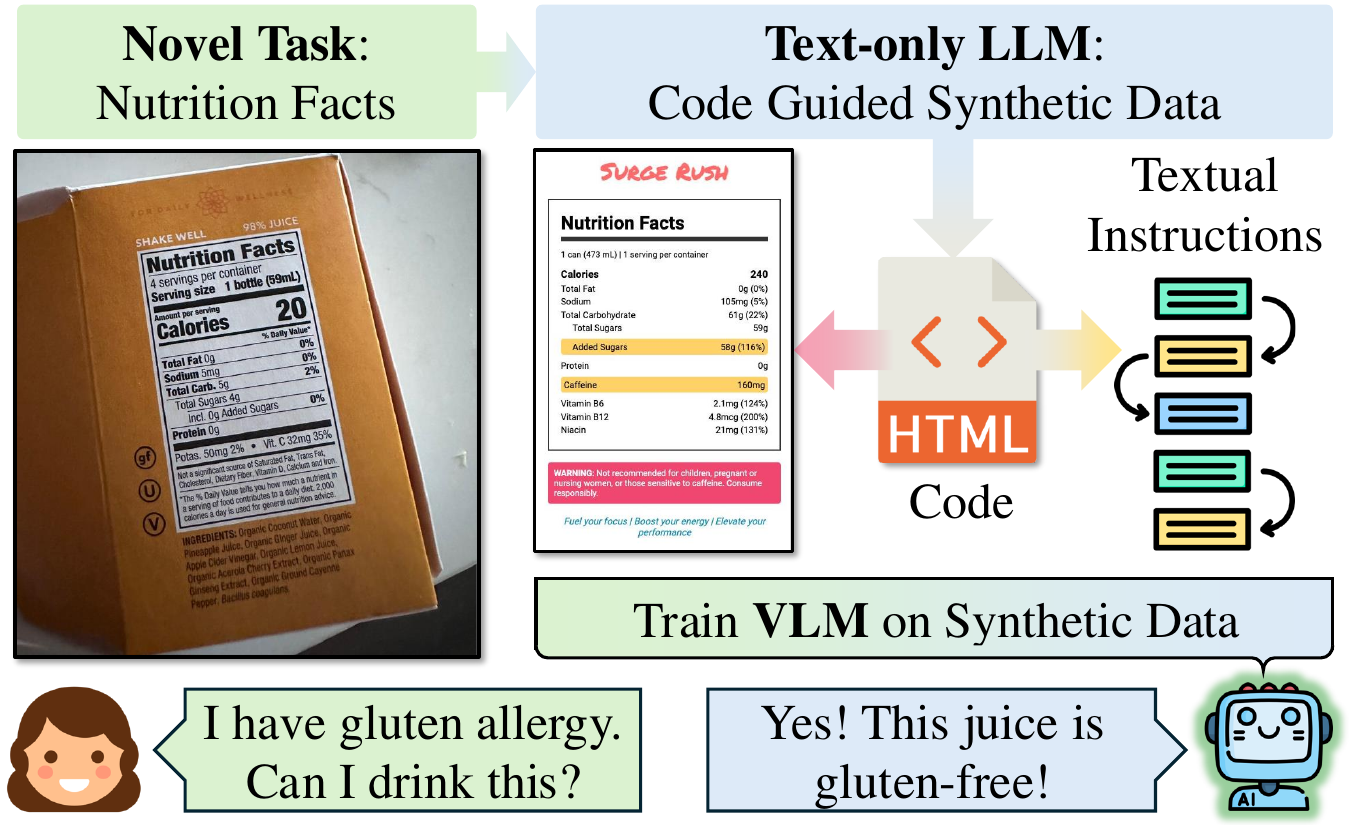}
    \vspace{-.6cm}
    \caption{Given a novel task (e.g., answering questions about nutrition facts), our code-guided generation system can produce targeted synthetic data to enhance the performance of VLMs on that specific task.}
    \label{fig: teaser}
    \vspace{-0.3cm}
\end{figure}

\section{Introduction}
Instruction-tuned vision-language models (VLMs) have shown strong performance across a range of multimodal tasks \citep{clip,gpt4,liu2023llava}.
However, these tasks typically focus on general image understanding over natural images rather than the specialized reasoning required for text-rich images such as charts, documents, diagrams, signs, labels, and screenshots.
Understanding and reasoning over text-rich images is crucial for many applications, including analyzing scientific literature and figures \cite{asai2024openscholar}, improving accessibility for users with visual impairments \cite{gurari2018vizwiz}, and enabling agentic workflows in real-world environments \cite{OSWorld}.
Effectively interpreting these structured visual formats requires both textual comprehension and spatial reasoning, which current models struggle with due to the limited availability of high-quality, realistic, and diverse vision-language datasets \citep{methani2020plotqa}.

\begin{figure*}[!t]
    \centering
    \includegraphics[width=\textwidth]{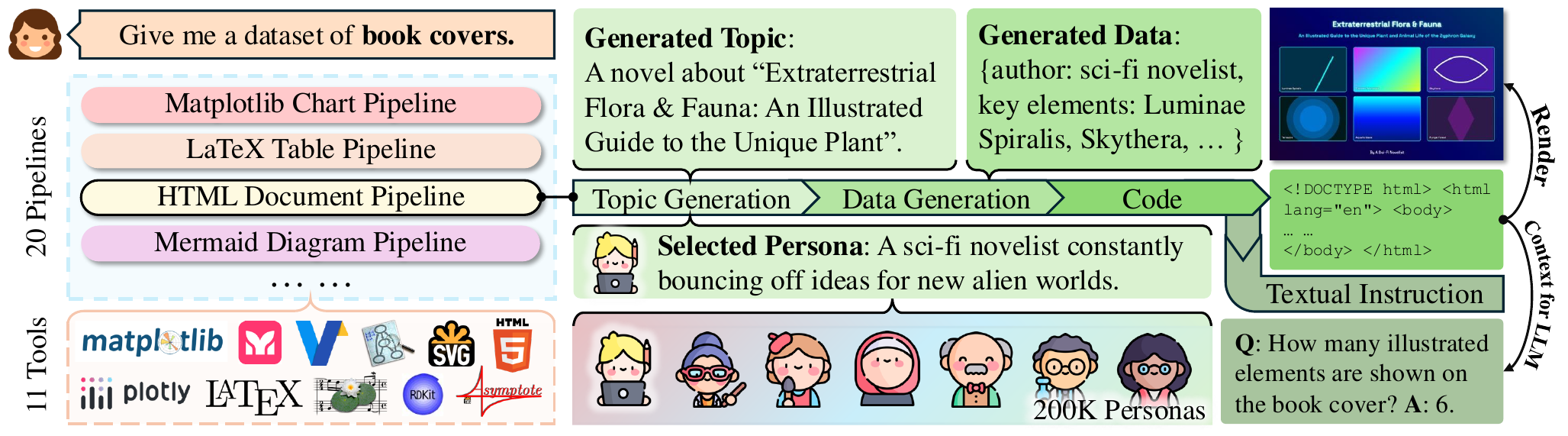}
    \vspace{-.6cm}
    \caption{The overview of our \textbf{Co}de Guided \textbf{Syn}thetic data generation system (\textbf{CoSyn}), which has 20 generation pipelines based on 11 render tools. Given a user query, e.g., ``book cover,'' CoSyn selects the appropriate pipelines and starts by generating diverse topics conditioned on personas, then synthesizes detailed data for code generation. The code renders the image and is also fed as context for an LLM to construct instruction-tuning data.}
    \label{fig: system}
    \vspace{-0.2cm}
\end{figure*}

To address these challenges and inspired by the fact that text-rich images are typically rendered from code, we develop \textbf{Co}de Guided \textbf{Syn}thetic data generation system (\textbf{CoSyn}), a flexible framework for generating diverse synthetic text-rich multimodal data for vision-language instruction tuning.
As illustrated in Figure \ref{fig: system}, CoSyn can generate multimodal data for various target domains from a short natural language query, such as \textit{book covers}.
CoSyn leverages text-only LLMs, which excel at code generation, to produce both data and code that render diverse text-rich images using 11 supported rendering tools (e.g., Python, HTML, LaTeX). 
Grounded in the underlying code representation of the images, textual instructions are also generated by the text-only LLM to create vision-language instruction-tuning datasets.

Using this framework, we construct the CoSyn-400K, as shown in Figure \ref{fig: dataset}, a large-scale and diverse synthetic vision-language instruction-tuning dataset tailored for text-rich image understanding.
We comprehensively evaluate the effectiveness of training on CoSyn-generated synthetic data across seven text-rich VQA benchmarks. 
Our model achieves state-of-the-art performance among competitive open-source models and surpasses proprietary models such as GPT-4V and Gemini 1.5.
Notably, training on CoSyn synthetic data enables sample-efficient learning, achieving stronger performance with less data.
In addition, CoSyn can synthesize chain-of-thought (CoT) reasoning data \citep{wei2022chain}, improving performance on tasks requiring multi-hop reasoning. 
A fine-grained analysis of question types in ChartQA \cite{masry-etal-2022-chartqa} reveals that training on CoSyn-400K results in stronger generalization to human-written questions. 
In contrast, models trained solely on existing academic datasets often overfit to biased training data, overperforming on templated or machine-generated questions but struggling with more realistic, human-asked queries.

We then identify a key limitation of open-source VLMs that they struggle to generalize to out-of-domain tasks they were not trained on.
As shown in Figure \ref{fig: teaser}, we introduce NutritionQA, a novel benchmark for understanding photos of nutrition labels, with practical applications like aiding users with visual impairments.
Open-source VLMs perform poorly on this novel task, even after training on millions of images. 
However, by training on CoSyn-400K, our model adapts strongly to this novel domain in a zero-shot setting with significantly less training data.
Remarkably, by generating just 7K in-domain synthetic nutrition label examples using CoSyn for fine-tuning, our model surpasses most open VLMs trained on millions of images. 
This highlights CoSyn’s data efficiency and ability to help VLMs adapt to new domains through targeted synthetic data generation.

Finally, beyond the standard VQA task, we use CoSyn to generate synthetic \textit{pointing} training data, which is particularly useful in agentic tasks. 
The pointing data enables VLMs to retrieve coordinates for specific elements in a screenshot given a query like ``Point to the Checkout button'' \citep{deitke2024molmo}.
Our model trained on synthetic pointing data achieves state-of-the-art performance on the ScreenSpot click prediction benchmark \citep{baechler2024screenai}.
Overall, our work demonstrates that synthetic data is a promising solution for advancing vision-language models in understanding text-rich images and unlocking their potential as multimodal digital assistants for real-world applications.

\section{Related Work}
\textbf{Vision Language Models.} \citet{tsimpoukelli2021multimodal} first demonstrate that pre-trained, frozen language models can be extended to process visual inputs.
Previous works fuse vision and language modalities using different strategies, such as cross-attention mechanisms \citep{alayrac2022flamingo} and Q-Former \citep{li2023blip}. 
More recent architectures have converged on using MLP layers to project visual features into the language space \citep{liu2023llava}.
However, these architectures are often imbalanced, with the language backbone substantially larger than the visual encoder. 
As a result, without high-quality image-text data, models may overly rely on language priors, leading to hallucinations in their responses \citep{bai2024hallucination}.
Our work addresses this issue by generating high-quality multimodal data for text-rich images.

\smallbreak
\noindent \textbf{Text-rich Images Understanding.} Chart understanding and text-rich image understanding continue to challenge state-of-the-art models as naturally occurring vision-language data that can support training for understanding text-rich images is still scarce \citep{kahou2017figureqa,kafle2018dvqa,xu2023chartbench,mukhopadhyay-etal-2024-unraveling}. In addition to charts and plots, a number of datasets address other kinds of text-rich images such as documents, infographics, diagrams, and figures, and screenshots \citep{figureseer,mathew2021docvqa,mathew2022infographicvqa,baechler2024screenai,roberts2024scifibench} have been made available. Many of these benchmarks are limited in size and scope, diversity of visualization types, and question types, making them suitable for evaluation but not for training data that could lead to generalized performance.

\smallbreak
\noindent \textbf{Synthetic Data for VLM.} Generating synthetic images with annotations grounded in known source representations has been widely used in domains with limited vision-language data \citep{virtualworldannotation,clevr,simvqa,provision}. 
This approach has been applied to chart and plot VQA, typically using a limited, small set of chart types and by instantiating handcrafted question templates \citep{kahou2017figureqa,kafle2018dvqa,methani2020plotqa,leafqa++}.
Following this, \citet{scigraphqa} and \citet{chartbasedreasoning} explore using text-only LLMs to generate annotations from tables or text descriptions associated with charts to train VLMs. 
Other recent approaches, similar to our procedure, explore generating data and code to render synthetic charts \cite{chartllama,he2024distill,sbsfigures,chartx}.
Molmo \cite{deitke2024molmo} releases a synthetic text-rich image dataset, \href{https://huggingface.co/datasets/allenai/pixmo-docs}{PixMo-docs}, but smaller in scale and diversity than ours.
These works generate synthetic data that is still highly limited in terms of the diversity of topics, figure types, and rendering pipelines, which is important for generalizing to out-of-distribution tasks. 
We expand the scope beyond charts to encompass a wider range of diverse text-rich images.

\section{Problem Formulation}
Given a text query $q$ about an image type, e.g., \textit{flow charts}, our goal is to create a synthetic multimodal dataset $\mathcal{D}_q = \left\{\left(I, T\right)\right\}$, where $I$ is the image, and $T$ is the textual instruction-tuning data (e.g., question-answer pairs). 
$\mathcal{D}_q$ is used to train a VLM to improve its ability to understand images related to $q$. 
The core idea of our approach is using code $C$ as the intermediate representation to bridge the image and text. 
The overall generation process can be decomposed as follows:
\begin{equation*}
    P\left(I, T | q\right) = P_{\text{LM}}\left(C|q\right) \cdot P\left(I|C\right) \cdot P_{\text{LM}}\left(T|C\right)
\end{equation*}
where $P_\text{LM}\left(C|q\right)$ represents prompting a language model to generate code $C$, which is executed to render the image, $P\left(I|C\right)$.
$P_\text{LM}\left(T|C\right)$ uses code $C$ (without the image) as context for an LLM to generate the textual instruction-tuning data.

\section{CoSyn System} \label{sec: cosyn}
Figure \ref{fig: system} illustrates the workflow of our \textbf{Co}de-Guided \textbf{Syn}thetic data generation system (\textbf{CoSyn}). 
The system takes a language input, such as ``generate a dataset of book covers'', and outputs a multimodal dataset. 
Based on the input query, CoSyn selects one of 20 generation pipelines built on 11 rendering tools.
The process starts with topic generation, conditioned on a sampled persona that guides the style and content. 
Next, the system generates data content and converts it into code, which is then executed to render synthetic images. 
Finally, using the code as context, we prompt the LLM to generate corresponding textual instructions.

In the following, we provide detailed explanations of the rendering tools supported by CoSyn, the tailored generation pipelines based on these tools, our persona-driven approach to diversifying content and styles, and the large-scale dataset of 400K synthetic images generated by CoSyn. 

\smallbreak
\noindent \textbf{Rendering Tools.} We integrate various rendering tools to generate diverse types of images, forming the foundation of CoSyn’s ability for text-rich image generation. 
For example, \href{https://matplotlib.org/}{Matplotlib}, \href{https://plotly.com/}{Plotly}, and \href{https://vega.github.io/vega-lite/}{Vega-Lite} are used to create different types of charts.
LaTeX and HTML are used for documents and tables, while \href{https://mermaid.js.org/}{Mermaid} and \href{https://graphviz.org/}{Graphviz} generate diagrams.
We utilize SVG and \href{https://asymptote.sourceforge.io/}{Asymptote} to create vector graphics and math-related content. 
For specialized tasks, we rely on \href{http://lilypond.org/}{Lilypond} to generate music sheets and \href{https://www.rdkit.org/}{RDKit} for chemical structures.
We implement customized functions for each tool to execute LLM-generated code and obtain corresponding rendered images. 
These tools collectively enable CoSyn to produce a wide range of high-quality, text-rich synthetic images.
\begin{figure*}[!t]
    \centering
    \includegraphics[width=\textwidth]{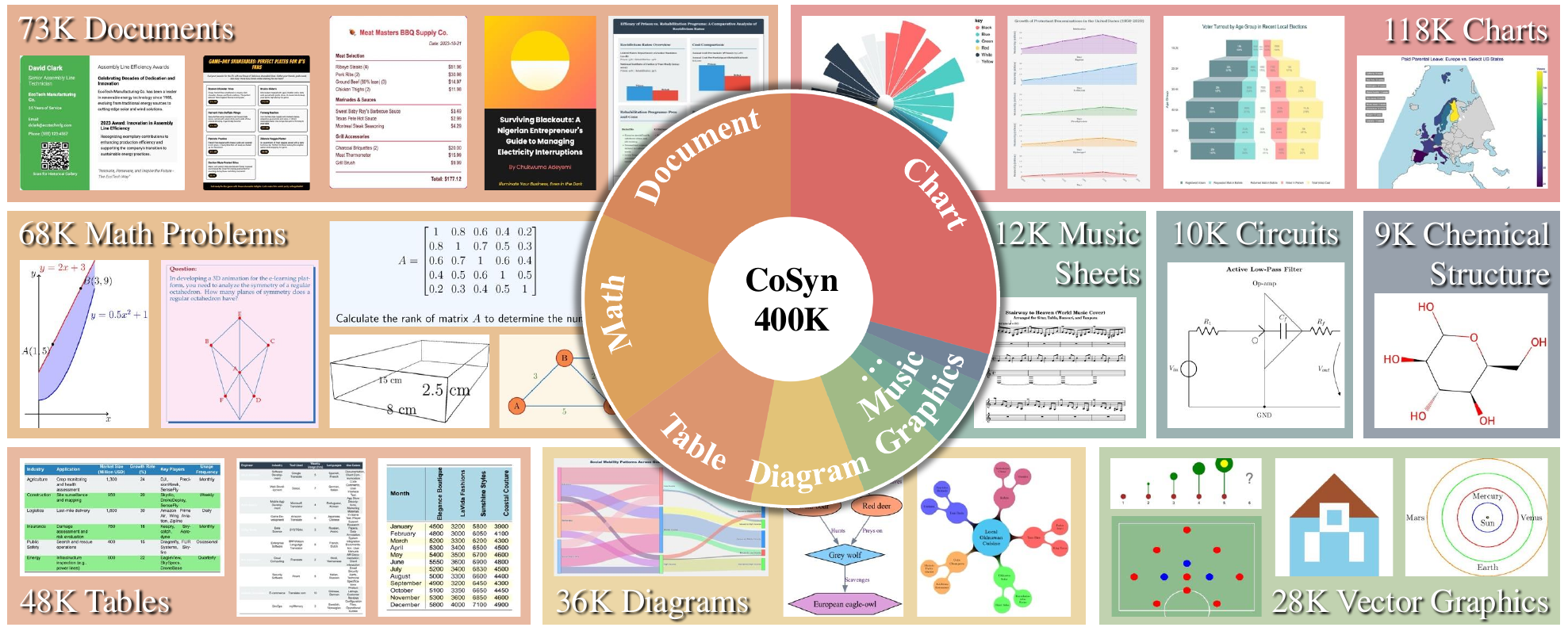}
    \vspace{-.6cm}
    \caption{Our CoSyn-400K dataset consists of 9 categories of text-rich images with 2.7M instruction-tuning data. More qualitative examples, along with question-answer annotations, are available in Figure \ref{fig: chart_example} -\ref{fig: special_example} in Appendix \ref{appendix: example}.}
    \label{fig: dataset}
    \vspace{-.3cm}
\end{figure*}

\smallbreak
\noindent \textbf{Pipelines.} We design 20 pipelines based on 11 rendering tools.\footnote {Some tools are used in multiple pipelines, e.g., HTML is used for generating documents, tables, and charts.} 
Each pipeline follows the same procedure: (1) \textit{Topic generation} to define the theme of this synthetic example, (2) \textit{Data generation} to populate the detailed contents, (3) \textit{Code generation} to create executable code that renders the image, and (4) \textit{Instruction generation} conditioned on code to produce instructions, including questions, answers and explanations for chain-of-thought reasoning.
Each stage is controlled by a prompt customized for the image category and rendering tool. Figure \ref{fig:prompt-html} shows prompts of the HTML Document pipeline.
\smallbreak
\noindent \textbf{Use personas to enhance diversity.} 
LLMs often struggle to generate diverse synthetic data using sampling parameters alone \citep{yu2023large}, with biases leading to repetitive outputs across different runs. 
Recent work \citep{ge2024scalingsyntheticdatacreation} shows that incorporating personas in prompts can improve diversity by enabling models to generate from varied perspectives.
CoSyn adopts personas to enhance diversity during the Topic Generation stage. 
Each persona is a short sentence describing a personality or identity. 
For example, as shown in the middle of Figure \ref{fig: system}, we sample a persona ``\textit{a sci-fi novelist who likes alien worlds}'', which results in a topic of ``\textit{a novel about Extraterrestrial Flora \& Fauna}'' for generating the book cover image. 
We use the 200K personas released by \citet{ge2024scalingsyntheticdatacreation}.

\smallbreak
\noindent \textbf{Implementation details.} 
CoSyn is built on the DataDreamer library \citep{patel-etal-2024-datadreamer}, which supports robust multi-stage synthetic data generation pipelines that are easy to maintain, reproduce, and extend. 
DataDreamer documents the prompts and parameters used at each generation stage and implements several efficient techniques, such as parallel generation and response caching, to optimize performance.
For the data and code generation stages, we use Claude-3.5-Sonnet, which performs well in coding tasks \citep{Anthropic}. 
For instruction-tuning data generation, we select GPT-4o-mini \citep{gpt4} for its cost efficiency. 

\smallbreak
\noindent \textbf{CoSyn-400K.} As shown in Figure \ref{fig: dataset}, we use CoSyn to generate a large-scale synthetic dataset of 400K images across nine categories: charts, documents, math problems, tables, diagrams, vector graphics, music sheets, electrical circuits, and chemical structures.
Since CoSyn is controlled via language inputs, it can easily generate diverse, fine-grained image types by varying the input queries. 
For instance, we use over 100 queries to generate document data covering \textit{receipts}, \textit{resumes}, \textit{meal plans}, etc. 
Some queries used for CoSyn-400K are provided in Appendix \ref{appendix: query}. 
This ensures that our dataset covers a broad range of domains.
The following sections validate how our synthetic datasets enhance the ability of VLMs to understand text-rich images.

\section{Experimental Setup}
\begin{table*}[!ht]
    \small
    \centering
    \setlength{\tabcolsep}{4.5pt}
    \begin{tabular}{lcccccccccc}
        \toprule
        \textbf{Model} & \textbf{ChartQA} & \textbf{DocVQA} & \textbf{InfoVQA} & \textbf{TableVQA} & \textbf{AI2D} & \textbf{TextVQA} & \textbf{ScreenQA} & \textbf{Average} \\
        \midrule
        \textcolor{gray}{GPT-4V} & \textcolor{gray}{78.1} & \textcolor{gray}{87.2} & \textcolor{gray}{75.1} & \textcolor{gray}{60.5} & \textcolor{gray}{89.4} & \textcolor{gray}{78.0} & \textcolor{gray}{41.6} & \textcolor{gray}{72.8} \\
        \textcolor{gray}{Gemini 1.5 Flash} & \textcolor{gray}{85.4} & \textcolor{gray}{89.9} & \textcolor{gray}{75.3} & \textcolor{gray}{72.6} & \textcolor{gray}{91.7} & \textcolor{gray}{78.7} & \textcolor{gray}{40.1} & \textcolor{gray}{76.2} \\
        \textcolor{gray}{Claude-3 Opus} & \textcolor{gray}{80.8} & \textcolor{gray}{89.3} & \textcolor{gray}{55.6} & \textcolor{gray}{70.0} & \textcolor{gray}{88.1} & \textcolor{gray}{67.5} & \textcolor{gray}{39.8} & \textcolor{gray}{70.2} \\ \midrule
        PaliGemma-3B$^\dagger$ & 71.4 & 84.8 & 47.8 & 46.4 & 73.3 & 76.5 & 32.2 & 61.8  \\
        BLIP-3-4B$^\dagger$ & 60.0 & 61.4 & 31.5 & 24.3 & 74.2 & 71.0 & 26.2 & 49.8 \\ 
        Cambrian-7B$^\dagger$ & 73.3 & 77.8 & 41.6 & 40.6 & 73.0 & 71.7 & 44.4 & 64.2 \\
        LLaVA-1.5-7B$^\dagger$$^*$ & 17.8 & 28.1 & 25.8 & 33.1 & 55.5 & 58.2 & 17.6 & 33.7 \\
        LLaVA-Next-8B$^\dagger$ & 69.5 & 78.2 & 43.8 & 43.9 & 71.6 & 65.3 & 34.2 &  58.1 \\ 
        LLaVA-OneVision-7B$^\dagger$ & 80.0 & 87.5 & \underline{68.8} & 64.6 & \underline{81.4} & \underline{78.3} & 46.3 & 72.4 \\
        Pixtral-12B & 81.8 & \textbf{90.7} & 50.8 & \textbf{67.0} & 79.0 & 75.7 & 39.4 & 69.2 \\ 
        Llama 3.2 11B & \underline{83.4} & 88.4 & 63.6 & 51.1 & \textbf{91.9} & 73.1 & \textbf{87.7} & \underline{77.0} \\ \midrule
        \cellcolor{gray!10}Ours (7B)$^\dagger$ & \cellcolor{gray!10}\textbf{86.3} & \cellcolor{gray!10}\underline{90.0} & \cellcolor{gray!10}\textbf{70.5} & \cellcolor{gray!10}\underline{65.8} & \cellcolor{gray!10}\textbf{91.9} &  \cellcolor{gray!10}\textbf{82.0} & \cellcolor{gray!10}\underline{80.1} & \cellcolor{gray!10}\textbf{80.9} \\ 
        \cellcolor{gray!10}Ours (7B-zero-shot)$^\dagger$$^*$ & \cellcolor{gray!10}80.8 & \cellcolor{gray!10}82.9 & \cellcolor{gray!10}59.8 & \cellcolor{gray!10}64.9 & \cellcolor{gray!10}83.9 & \cellcolor{gray!10}72.7 & \cellcolor{gray!10}78.1 & \cellcolor{gray!10}74.7  \\
        \bottomrule
    \end{tabular}
    \vspace{-.1cm}
    \caption{\textbf{Results on 7 text-rich benchmarks.} The result of the best-performing open-source model is \textbf{bold}, and the second-best is \underline{underlined}. Models with $^\dagger$ stand for open data and code for multimodal training. Models with $^*$ are zero-shot models, which means the models are not trained on instances from any of the evaluation datasets.}
    \label{tab:model_performance}
    \vspace{-.3cm}
\end{table*}

Our experiments aim to verify the value of our synthetic data in the supervised fine-tuning stage of training vision-language models.
This section introduces the architecture of our model, training strategy, datasets we used, baselines for comparison, and other details on implementation.
\smallbreak
\noindent \textbf{Model Architecture.} We follow the same image preprocessing and architecture as Molmo \cite{deitke2024molmo}, which uses the MLP layer to connect the vision encoder and a pretrained LLM. We choose OpenAI’s CLIP (ViT-L/14 336px) \cite{clip} as the vision backbone and Mistral-7B \cite{jiang2023mistral} as the language model.
\smallbreak
\noindent \textbf{Training Process.} We adopt the same training strategy as Molmo \cite{deitke2024molmo}, which consists of two stages: (1) \textit{Pre-training} on dense captions from \href{https://huggingface.co/datasets/allenai/pixmo-cap}{PixMo-Cap} and (2) \textit{Supervised fine-tuning} on three categories of datasets below:

\begin{itemize}[noitemsep, topsep=0pt, leftmargin=*]
    \item \textbf{Evaluation Datasets.} We evaluate our model on seven text-rich benchmarks, including ChartQA \cite{masry-etal-2022-chartqa}, DocVQA \cite{mathew2021docvqa}, InfographicVQA \cite{mathew2022infographicvqa}, TableVQA-Bench \cite{kim2024tablevqa}, AI2 Diagrams \cite{kembhavi2016diagram}, TextVQA \cite{singh2019towards}, and ScreenQA \cite{baechler2024screenai}. We adopt their official metrics for calculating performance. In total, we have 138K training images from the evaluation datasets.\footnote{TableVQA is an eval-only benchmark (no training split), and we do not use the training split from ScreenQA.}

    \item \textbf{Auxiliary Datasets.} We select additional academic datasets for fine-tuning: VQAv2 \cite{balanced_vqa_v2}, GQA \cite{hudson2019gqa}, OK-VQA \cite{okvqa}, OCR-VQA \cite{mishraICDAR19}, A-OKVQA \cite{schwenk2022okvqa}, ScienceQA \cite{lu2022learn}, TabMWP \cite{lu2023dynamic}, ST-VQA \cite{biten2019scene}, TallyQA \cite{acharya2019tallyqa}, DVQA \cite{kafle2018dvqa}, FigureQA \cite{kahou2017figureqa}, and PlotQA \cite{methani2020plotqa}. The auxiliary datasets contain around 1M training images.

    \item \textbf{Synthetic Datasets.} As introduced in Sec \ref{sec: cosyn} and also shown in Figure \ref{fig: dataset}, our synthetic datasets include 400K text-rich images from 9 categories.
\end{itemize}

\noindent Our best-performing model uses all three categories of datasets above. We also trained a zero-shot model using only auxiliary and synthetic data without any examples from the evaluation datasets, which still exhibits competitive benchmark performance, as shown in the last row of Table \ref{tab:model_performance}.
\smallbreak
\noindent \textbf{Baselines.} We compare recent open-source VLMs with a similar scale (7B), including PaliGemma-3B \cite{beyer2024paligemma}, BLIP-3-4B \cite{xue2024xgen}, Cambrian-7B \cite{tong2024cambrian}, LLaVA-1.5-7B \cite{liu2023llava}, LLaVA-Next-8B \cite{liu2024llavanext}, LLaVA OneVision-7B \cite{li2024llava}, Pixtral-12B \cite{agrawal2024pixtral}, Llama 3.2 V \cite{meta2024llama}.
We also include proprietary models: GPT-4V \cite{gpt4}, Gemini-1.5-Flash \cite{team2024gemini}, and Claude-3 Opus \cite{anthropic2024claude}.
\smallbreak
\noindent \textbf{Implementation Details.} We train our model on TPU v3-128 with a batch size of 32. Our best-performing model is trained for 60K steps, taking about 30 hours. The checkpoints with the highest validation performance are retained for testing.

\section{Results}

This section covers (1) the competitive performance of the model trained on our synthetic data (Sec \ref{sec: main_results}), (2) the comprehensive analyses to highlight the benefits of synthetic data (Sec \ref{sec: analysis}), and (3) the effectiveness of synthetic pointing data in improving VLMs for web agent tasks (Sec \ref{sec: pointing}).

\begin{figure}[!t]
    \centering
    \includegraphics[width=\columnwidth]{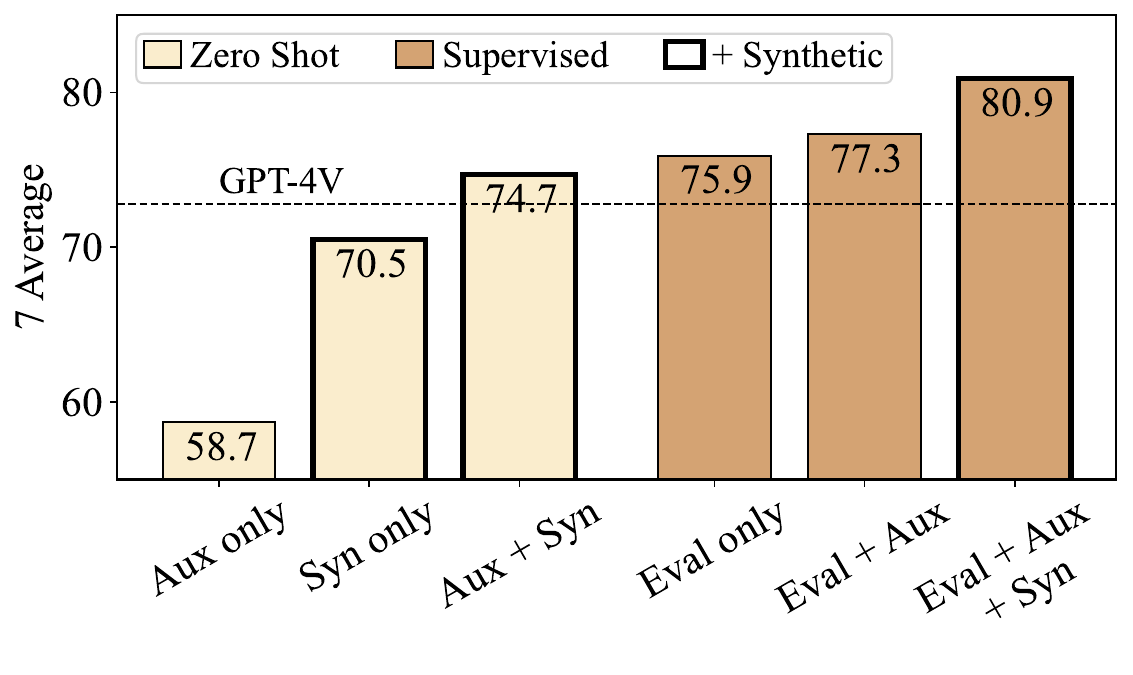}
    \vspace{-.8cm}
    \caption{\textbf{Ablation on training data selection.} Aux, Syn, and Eval stand for auxiliary, synthetic, and evaluation datasets, respectively. We report the average score on seven benchmarks. The detailed performance breakdown on each benchmark is in Table \ref{tab:train_data_ablation}.}
    \label{fig: ablation}
    \vspace{-0.3cm}
\end{figure}

\subsection{Main Results} \label{sec: main_results}

Table \ref{tab:model_performance} compares our model’s performance with both open and closed models across seven text-rich benchmarks. 
On average, our 7B model achieves the highest performance, surpassing the second-best model (Llama 3.2 11B) by 3.9\%. 
Notably, our model ranks first in four out of the seven datasets and second in the remaining three.
More surprisingly, our zero-shot model (the last row in Table \ref{tab:model_performance}) outperforms most open and closed models without exposure to any training instances from the evaluation datasets. 
In contrast, these competing models often rely on benchmark training data and are thus not true zero-shot models. 
This result demonstrates that the capabilities learned from our synthetic data can transfer effectively to downstream tasks.

\subsection{Analysis} \label{sec: analysis}
In the following experiments, we quantify the contribution of synthetic data to the benchmark performance by ablating the combinations of fine-tuning datasets. 
Then, we demonstrate that our CoSyn system can efficiently assist VLMs in generalizing to novel tasks. 
Finally, we show that synthetic data can help mitigate the overfitting of biases.
\smallbreak
\noindent \textbf{Synthetic data boosts the performance.} Figure \ref{fig: ablation} presents an ablation study on the choices of supervised fine-tuning data. 
In the zero-shot settings, when the model is trained on auxiliary datasets (over 1M training images not directly from the evaluation tasks), it fails to generalize effectively to the evaluation tasks, with a substantial performance gap of 14.1\% below GPT-4V.
However, using only 400K synthetic samples achieves a performance comparable to GPT-4V. 
Our best zero-shot model surpasses GPT-4V when jointly training synthetic and auxiliary data.
Under the supervised settings, training with in-domain data alone yields strong performance. 
However, adding 1M auxiliary samples provides a modest improvement of 1.4\%, while incorporating synthetic data results in a more significant 3.6\% boost. 
These results demonstrate the effectiveness of synthetic data in enhancing VLMs' performance on text-rich tasks.

\begin{figure}[!t]
    \centering
    \includegraphics[width=\columnwidth]{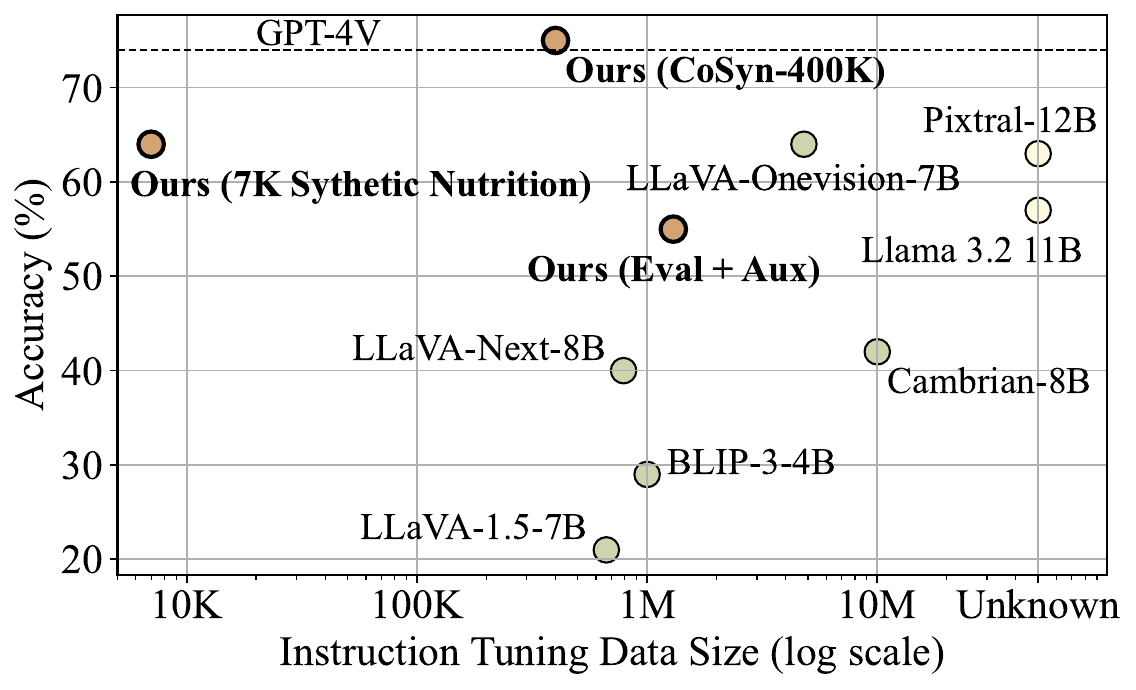}
    \vspace{-.6cm}
    \caption{\textbf{Zero shot performance on NutritionQA.} The x-axis denotes the number of training examples used for the instruction-tuning stage. The models on the upper left side demonstrate better data efficiency.}
    \label{fig: nutrition}
    \vspace{-0.3cm}
\end{figure}

\smallbreak
\noindent \textbf{Zero-shot Generalization on a Novel Task.} Vision-language models typically rely on in-domain data to perform well on specific tasks. 
When encountering a novel task, such as answering questions about nutrition labels in Figure \ref{fig: teaser}, models without seeing similar examples during training may struggle with this novel task. 
However, our CoSyn system enables controllable data generation. 
Given the task name as input, CoSyn can generate task-specific data to fine-tune the model.

To validate this, we annotated a small 
evaluation dataset called \href{https://huggingface.co/datasets/yyupenn/NutritionQA}{NutritionQA}, which includes 100 examples of questions about photos of nutrition labels. 
Some questions require multi-hop reasoning, as Figure \ref{fig: nurition_qa} illustrates. 
We evaluated GPT-4V and several open-source VLMs on this dataset and report the performance in Figure \ref{fig: nutrition}.
The x-axis in Figure \ref{fig: nutrition} represents the amount of data used during the instruction fine-tuning stage.

Despite being trained on millions of images, we observe that open-source VLMs are not data-efficient and perform poorly on this novel task compared to GPT-4V.
Although many open-source VLMs claim to achieve GPT-4V-level performance, they fall short when tested on new tasks in the wild. 
Without synthetic data, our model (Eval + Aux) achieves results similar to those of open models. 
However, when trained on 400K synthetic samples, our model matches GPT-4V’s performance.

More impressively, we used CoSyn to generate 7K synthetic nutrition label samples and fine-tuned the model using only this 7K data. 
The resulting model outperforms most open-source VLMs on the NutritionQA task. 
These results demonstrate that code-guided synthetic data is an effective and efficient method for adapting VLMs to new domains.

\begin{figure}[!t]
    \centering
    \includegraphics[width=\columnwidth]{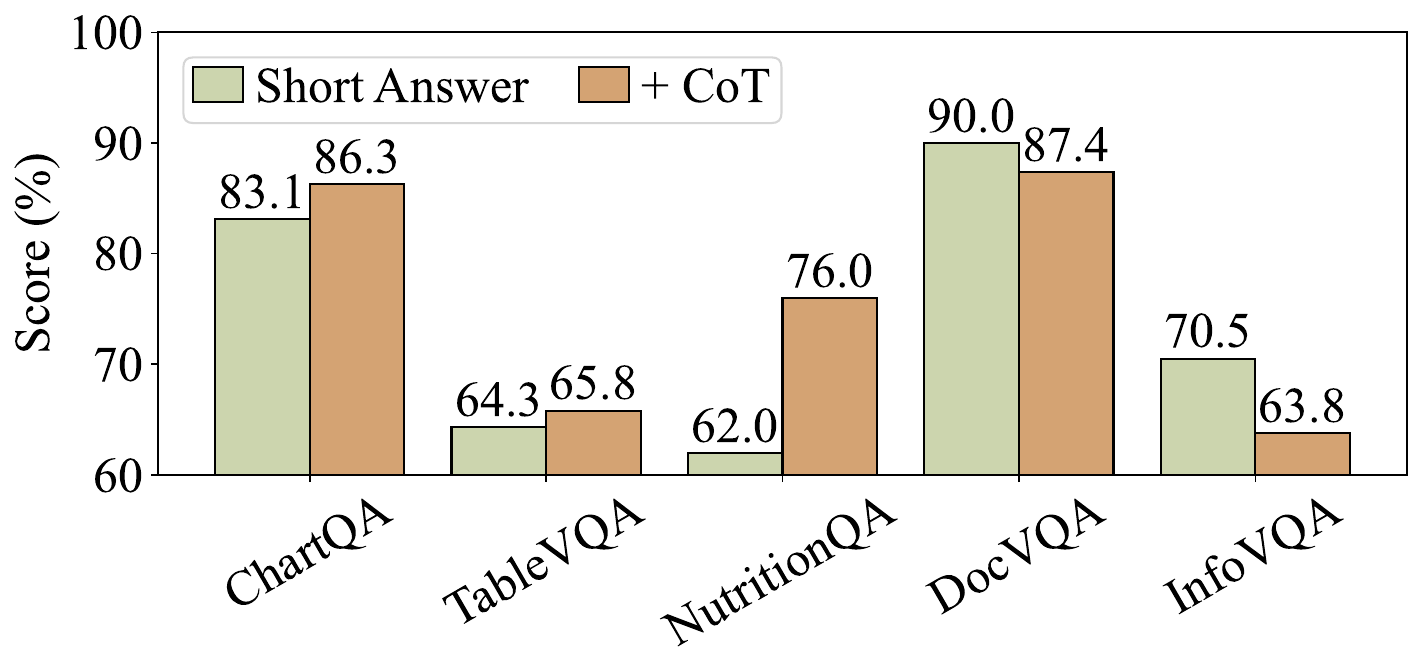}
    \vspace{-.6cm}
    \caption{\textbf{Ablation of using Chain-of-Thought reasoning.} Short Answer represents prompting model to output the answer as short as possible. $+$ CoT stands for providing Chain-of-Thought reasoning before giving the final answer. Results on all datasets are in Table \ref{tab:cot}.}
    \label{fig: cot}
    \vspace{-0.2cm}
\end{figure}

\smallbreak
\noindent \textbf{Synthetic Data for Chain-of-Thought Reasoning.} Existing text-rich datasets, such as ChartQA \cite{masry-etal-2022-chartqa}, are typically annotated with short answers. 
However, questions like ``Compute the mean of the data in the plot'' require step-by-step mathematical reasoning to arrive at the correct answer. 
Models trained only with short-answer supervision may fail to learn proper plot comprehension, but instead overfit to annotation biases in these datasets. 
On the contrary, our CoSyn-400K includes explanation text alongside the short answer. 
Each instruction-tuning example consists of a (\textit{question}, \textit{explanation}, \textit{short answer}) triplet, enabling models to learn chain-of-thought (CoT) reasoning.
During fine-tuning, we design two prompt templates for our synthetic data:
\begin{center} 
\begin{tcolorbox}[top=2pt, bottom=2pt, width=\linewidth, boxrule=1pt]
{\small {\fontfamily{put}\selectfont
\textbf{CoT Prompt}: \textbf{<Question>} Provide reasoning steps and then give the short answer.

\textbf{<Explanation>}
Answer: \textbf{<Answer>}
}}
\end{tcolorbox}

\begin{tcolorbox}[top=2pt, bottom=2pt, width=\linewidth, boxrule=1pt]
{\small {\fontfamily{put}\selectfont
\textbf{Short Answer Prompt}: \textbf{<Question>} Answer with as few words as possible. \textbf{<Answer>}
}}
\end{tcolorbox}
\end{center}

Those prompts allow VLMs to switch between the two answering styles and perform CoT reasoning when necessary. Figure \ref{fig: cot} shows that incorporating CoT reasoning improves performance on ChartQA, TableVQA, and NutritionQA, as these datasets contain examples requiring multi-hop reasoning.
However, we observe that adding CoT reasoning reduces performance on DocVQA and InfoVQA. 
We find that answer biases in these benchmarks cause this decline. 
Specifically, the ground-truth answers favor short responses, often penalizing more detailed and verbal responses. 
For instance, in DocVQA, the ground-truth for an example is ``T-Th'', whereas the model responds with ``Tuesday to Thursday''. 
Although the response is correct, the strict string-matching metric assigns it a zero score.
This highlights key limitations of current multimodal benchmarks, including answering biases and rigid evaluation metrics that fail to capture the full extent of a model’s capabilities.

\begin{table}[!t]
    \small
    \centering
    \includegraphics[width=0.45\textwidth]{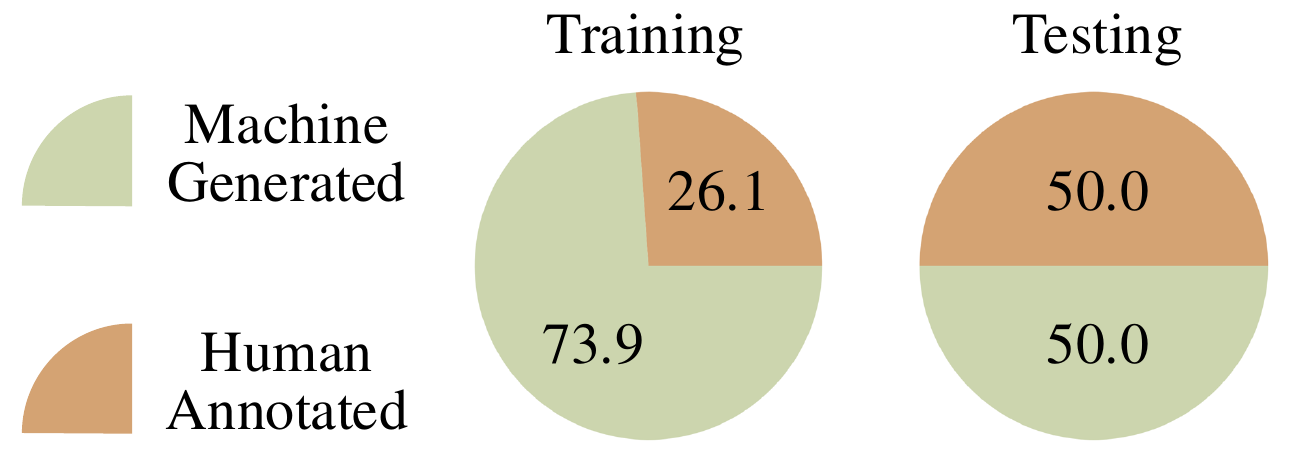} 
    \vspace{.3cm} 
    
    \setlength{\tabcolsep}{4pt}
    \begin{tabular}{lcccc}
        \toprule
        \textbf{ChartQA} & \textbf{Average} & \textbf{Machine} & \textbf{Human} &   \textbf{$\Delta \downarrow$} \\
        \midrule
        PaliGemma-3B & 71.4 & 88.5 & 54.2 &  34.3 \\
        ChartPali-5B & 77.3 & \textbf{93.7} & 60.9 &  32.8 \\ \midrule
        Ours (w/o Syn) & 81.4 & 92.2 & 70.4 &  21.8\\
        \cellcolor{gray!10}Ours (w/ Syn) & \cellcolor{gray!10}\textbf{86.3} & \cellcolor{gray!10}93.4 & \cellcolor{gray!10}\textbf{79.1} & \cellcolor{gray!10}\textbf{14.2}\\
        \bottomrule
    \end{tabular}
    \vspace{-.1cm}
    \caption{\textbf{Results on human and machine-generated questions of ChartQA.} 
    The pie charts above display the percentage distribution of two question types in training and testing. 
    $\Delta$ ($\downarrow$ lower is better) denotes the performance gap between human and machine questions.}
    \label{tab:chartqa}
    \vspace{-.3cm}
\end{table}
\smallbreak
\noindent \textbf{Synthetic Data for Mitigating Biases.} Our previous experiments reveal answering biases in multimodal benchmarks, which VLMs trained solely on these datasets often inherit. To further validate this issue, we analyze ChartQA and observe a distribution shift in question types. 
As shown in the pie charts above Table \ref{tab:chartqa}, some ChartQA questions are human-annotated, while others are generated by the language model T5 \citep{raffel2020exploring}, which is heavily influenced by prompt phrasing and limited to a fixed set of question templates.
During training, most questions (73.9\%) in ChartQA are machine-generated, while the test set contains an even distribution of human-annotated and machine-generated questions. Models trained exclusively on ChartQA tend to overfit to T5-generated questions. Table \ref{tab:chartqa} illustrates this issue: PaliGemma \citep{beyer2024paligemma} and ChartPali \citep{carbune2024chart} achieve high accuracy on machine-generated questions but experience a significant performance drop of over 30\% on human-annotated questions.

Similarly, without synthetic data, our model shows a noticeable 21.8\% gap between the two question types. 
However, incorporating synthetic data during training reduces this gap to 14.2\%. This suggests that synthetic data can mitigate overfitting on benchmarks and enhance VLMs' usability in real-world human-asked questions.

\smallbreak
\noindent \textbf{Our synthetic data is more diverse.} To quantify the diversity of images and text in our synthetic dataset $\mathcal{D} = \left\{\left(I, T\right)\right\}$, we propose the following two metrics to compute the diversity:
\vspace{-.1cm}
\begin{equation*}
\scalebox{0.78}{$
    \text{\textbf{Diversity}}(\mathcal{D})_\text{\textbf{Image}} = \frac{1}{|\mathcal{D}|^2-|\mathcal{D}|} \sum_{I_i \in \mathcal{D}} \sum^{i \neq j}_{I_j \in \mathcal{D}} \left(1 - \text{sim}(I_i, I_j)\right)
$}
\end{equation*}
\begin{equation*}
\scalebox{0.78}{$
    \text{\textbf{Diversity}}(\mathcal{D})_\text{\textbf{Text}} = \frac{1}{|\mathcal{D}|^2-|\mathcal{D}|} \sum_{T_i \in \mathcal{D}} \sum^{i \neq j}_{T_j \in \mathcal{D}} \left(1 - \text{sim}(T_i, T_j)\right)
$}
\vspace{.2cm}
\end{equation*}
where sim$(\cdot)$ is the cosine similarity function. Both metrics compute the average pairwise cosine distance between the features of every instance in the dataset.
For image diversity, we extract features using CLIP, while for text diversity, we use Sentence-BERT \citep{reimers2019sentence} to obtain embeddings of question-answer pairs.
Table \ref{tab:diversity} shows that our synthetic charts are significantly more diverse than those in existing datasets, such as FigureQA and ChartQA, in both image and text diversity.


\begin{table}[!t]
\centering
\small
\begin{tabular}{lcc}
\toprule
\textbf{Dataset} & \textbf{Image Diversity} & \textbf{Text Diversity} \\
\midrule
FigureQA & 0.268 & 0.567 \\
DVQA & 0.307 & 0.752 \\
PlotQA & 0.420 & 0.743 \\
ChartQA & 0.340 & 0.742 \\
\cellcolor[gray]{0.9}Ours (Charts) & \cellcolor[gray]{0.9}\textbf{0.596} & \cellcolor[gray]{0.9}\textbf{0.823} \\
\bottomrule
\end{tabular}
\vspace{-.1cm}
\caption{\textbf{Compare image and text diversity across different chart datasets.} We randomly sample 10K instances from each dataset to compute the results.}
\label{tab:diversity}
\end{table}

\begin{table}[!t]
    \small
    \centering
    \setlength{\tabcolsep}{4pt}
    \begin{tabular}{lccccc}
        \toprule
        \multirow{2}{*}{\textbf{n. of Tools}} & \multirow{2}{*}{\textbf{Diversity}} & \multicolumn{3}{c}{\textbf{ChartQA}} \\ \cmidrule{3-5}
        & & Average & Machine & Human \\
        \midrule
        \textsc{Single} & 0.572 & 73.9 & 66.5 & 81.5 \\ 
        \textsc{Multiple} & \textbf{0.607} & \textbf{75.2} & \textbf{68.6} & \textbf{82.0}\\
        \bottomrule
    \end{tabular}
    \vspace{-.1cm}
    \caption{\textbf{Single vs. Multiple Rendering Tools for Data Generation.} Each row uses the same number of 45K synthetic images. \textsc{Single} only uses Matplotlib, while \textsc{Multiple} involves four other rendering tools: HTML, LaTeX, Plotly, and VegaLite. }
    \label{tab:diversity_performance}
    \vspace{-.3cm}
\end{table}

\smallbreak
\noindent \textbf{Diversity correlates with model performance.} We observe that data diversity significantly affects model performance on downstream tasks. 
To investigate this, we compare synthetic chart data generated using only a single tool (Matplotlib) with charts generated by all five tools available in our CoSyn system. 
As shown in Table \ref{tab:diversity_performance}, using multiple tools results in higher image diversity and notably improved performance on ChartQA. This experiment underscores the importance of data diversity for enhancing the generalizability of models.

\begin{figure}[!t]
  \begin{center}
    \includegraphics[width=7.5cm]{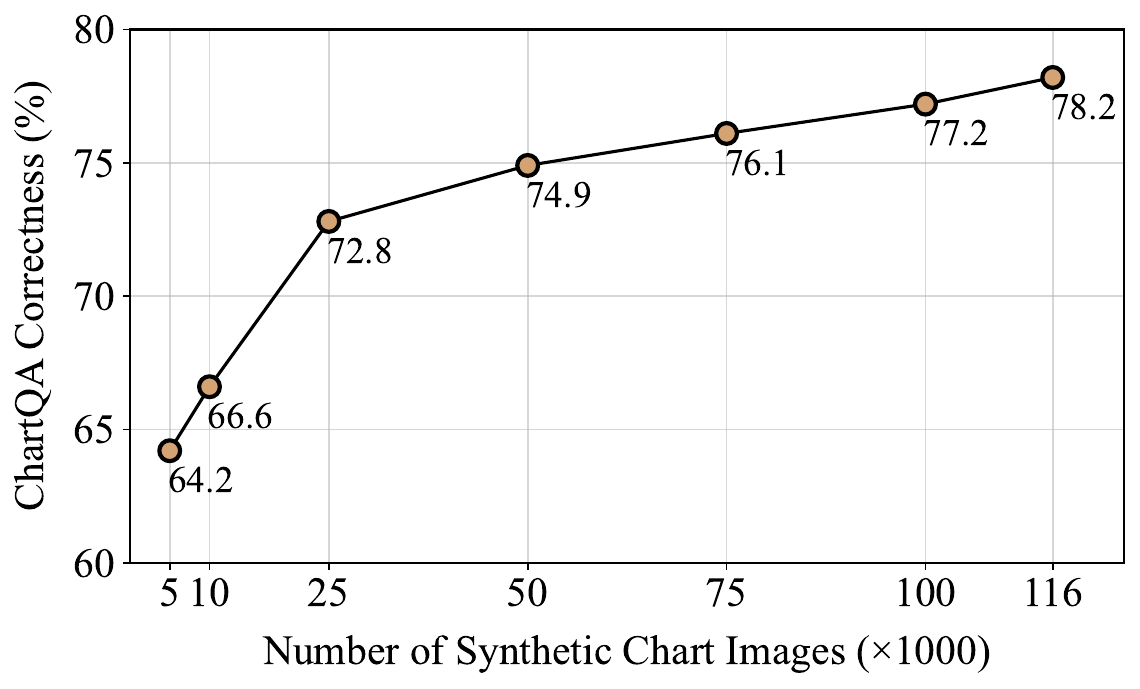}
  \end{center}
  \vspace{-.3cm}
  \caption{\textbf{Scaling the Size of Synthetic Data.} We evaluate the \textit{zero-shot} performance on ChartQA of models fine-tuned on increasing numbers of synthetic images.}
\label{fig: scale}
\end{figure}

\smallbreak
\noindent \textbf{Scaling the size of synthetic data.} In addition to diversity, the scale of synthetic data also impacts model performance. 
As shown in Figure \ref{fig: scale}, increasing the number of synthetic chart images leads to improved performance on ChartQA. This demonstrates that scaling up synthetic data can further enhance VLMs on downstream tasks. 
Due to resource constraints, our final dataset consists of 400K images, which cost us about \$8,000.
Future work could explore scaling up the dataset size to push the boundaries of synthetic data's potential.

\smallbreak
\begin{table}[!t]
    \small
    \centering
    \setlength{\tabcolsep}{4pt}
    \begin{tabular}{lccc}
        \toprule
        \multirow{2}{*}{\textbf{LLM for Data Generation}} & \multicolumn{3}{c}{\textbf{ChartQA}} \\ \cmidrule{2-4}
        & Average & Machine & Human \\
        \midrule
        GPT-4o & 72.4 & 65.8 & 78.9 \\ 
        Claude-3.5-sonnet & \textbf{77.2} & \textbf{71.0} & \textbf{83.8} \\
        \bottomrule
    \end{tabular}
    \vspace{-.1cm}
    \caption{\textbf{Compare the LLMs used for synthetic data generation.} For both LLMs, we create 100K synthetic charts for fine-tuning the VLMs. We report the zero-shot evaluation results on ChartQA.}
    \vspace{-.3cm}
    \label{tab:llm_ablate}
\end{table}

\noindent \textbf{Compare LLMs for synthetic data generation.} In the default setting, CoSyn uses Claude-3.5-sonnet as the underlying LLM for code generation. 
\noindent To highlight the importance of strong coding capabilities, we compare it with data generated by GPT-4o. 
As shown in Table \ref{tab:llm_ablate}, synthetic data generated by Claude-3.5-sonnet yields significantly better results than GPT-4o.
Our qualitative observation reveals that GPT-4o has a higher failure rate in code generation, particularly for less common coding languages or libraries. 
This result emphasizes that a strong LLM is essential for the successful synthetic data generation for VLMs.

\begin{figure*}[!t]
    \centering
    \includegraphics[width=\textwidth]{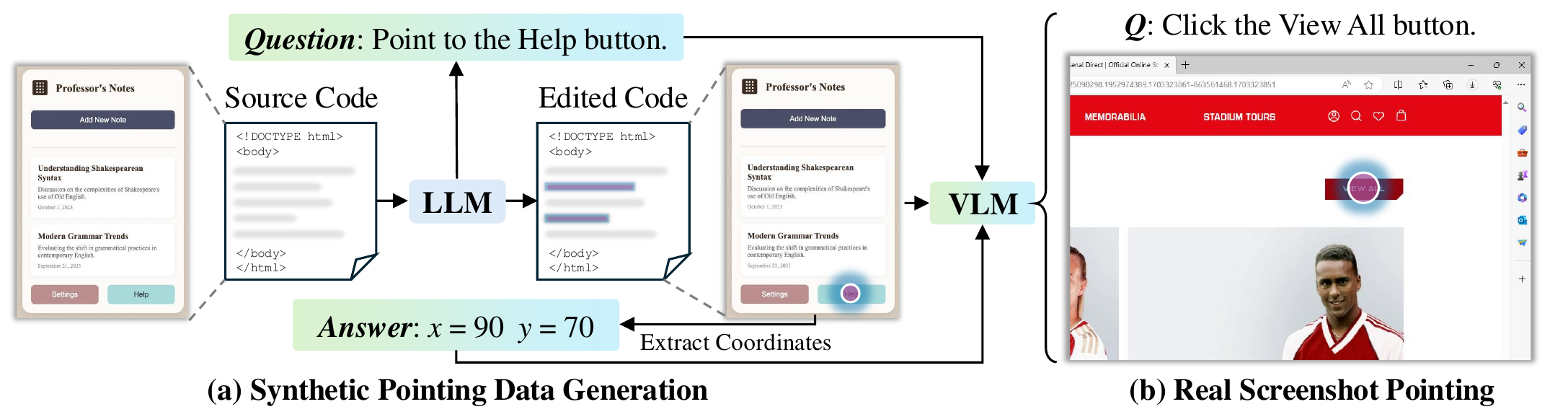}
    \vspace{-.5cm}
    \caption{\textbf{The overview of enabling VLMs to point through synthetic data.} (a) We synthesize pointing data by prompting an LLM to generate pointing questions and edit the code to draw the answer points explicitly. (b) We demonstrate that the VLM trained on synthetic pointing data can be generalized to real agentic tasks.}
    \label{fig: pointing}
    \vspace{-.2cm}
\end{figure*}

\subsection{Synthetic Pointing Data} \label{sec: pointing}
Pointing enables VLMs to answer questions by providing specific points on images. This allows models to ground their responses in visual content and interact with environments, which is crucial for developing digital agents.
We find that our code-guided system can synthesize pointing data.

\smallbreak
\noindent \textbf{Method.}
Since we have access to the source code for all generated images, we can prompt an LLM to modify the code to draw points on the images explicitly. 
As illustrated in Figure \ref{fig: pointing}, we feed the image’s source code as context to the LLM, which generates a pointing question and edits the code to draw points with a predefined color. 
By extracting the pixel values of these points, we can obtain their exact $(x, y)$ coordinates.\footnote{The coordinates of points are normalized to (0, 100) to mitigate the influence of image resolution.} 
We then use this pointing data to train VLMs, enabling them to answer questions by providing point coordinates. 
In total, we generate pointing data for 65K synthetic images. 
Figure \ref{fig: point_example} shows some qualitative examples from our synthetic pointing dataset.

\smallbreak
\noindent \textbf{Setup.}
We evaluate pointing ability on ScreenSpot \cite{cheng2024seeclick}, where the task requires models to provide the correct click location based on a given instruction. ScreenSpot contains screenshots from mobile phones, desktops, and web pages. 
To assess the effectiveness of our synthetic pointing data, we compare it to the model trained on PixMo-point \cite{deitke2024molmo}, which consists of 155K human-annotated images. 
Our best-performing model uses both PixMo-point and synthetic pointing data.
Additionally, we compare against existing methods like CogAgent \cite{hong2024cogagent}, SeeClick \cite{cheng2024seeclick}, and UGround \cite{gou2024navigating}, which is trained on 1.3M screenshots.

\smallbreak
\noindent \textbf{Results.}
Table \ref{tab:click} compares the click accuracy of our models with previous methods.
Using 65K synthetic pointing samples, our model achieves performance comparable to that trained on 155K human-annotated samples.
When combining synthetic and human data, our model achieves state-of-the-art performance on ScreenSpot, surpassing the recent UGround \cite{gou2024navigating}, which was trained on 1.3M screenshots. These results demonstrate that synthetic pointing data is a data-efficient approach for improving VLM performance on agentic tasks involving click prediction.

\begin{table}[!t]
    \centering
    \small
    \setlength{\tabcolsep}{4pt}
    \begin{tabular}{lcccccccc}
        \toprule
        & \multicolumn{2}{c}{\textbf{Mobile}} & \multicolumn{2}{c}{\textbf{Desktop}} & \multicolumn{2}{c}{\textbf{Web}} & \textbf{Avg} \\
        \cmidrule(lr){2-3} \cmidrule(lr){4-5} \cmidrule(lr){6-7}
        \textbf{Model} & Text & Icon & Text & Icon & Text & Icon & \\
        \midrule
        GPT-4o & 20.2 & 24.9 & 21.1 & 23.6 & 12.2 & 7.8 & 18.3 \\
        CogAgent & 67.0 & 24.0 & 74.2 & 20.0 & 70.4 & 28.6 & 47.4 \\
        SeeClick & 78.0 & 52.0 & 72.2 & 30.0 & 55.7 & 32.5 & 53.4 \\
        UGround & 82.8 & \underline{60.3} & 82.5 & \underline{63.6} & \underline{80.4} & \textbf{70.4} & \underline{73.3} \\ \midrule
        Synthetic & \textbf{90.8} & 53.3 & 78.4 & 58.6 & 80.0 & 47.1 & 68.0 \\
        Human & 84.2 & 59.0 & \textbf{88.1} & 52.9 & 76.5 & 50.5 & 68.5 \\
        \cellcolor{gray!10}Combined & \cellcolor{gray!10}\underline{89.0} & \cellcolor{gray!10}\textbf{65.1} & \cellcolor{gray!10}\underline{87.6} & \cellcolor{gray!10}\textbf{65.7} & \cellcolor{gray!10}\textbf{83.0} & \cellcolor{gray!10}\underline{58.7} & \cellcolor{gray!10}\textbf{74.9} \\
        \bottomrule
    \end{tabular}
    \caption{\textbf{Click accuracy on ScreenSpot.} We report our models trained on different pointing data. Human stands for using the human-annotated data from PixMo-point \cite{deitke2024molmo}. Combined means combining human-annotated data with our synthetic pointing data.}
    \label{tab:click}
    \vspace{-.2cm}
\end{table}

\section{Conclusion}

In this work, we introduced CoSyn, a framework for generating synthetic data that significantly enhances VLM performance on text-rich image understanding.
Our comprehensive analysis highlights the advantages of synthetic data for domain generalization, data efficiency, and bias mitigation. 
Our work demonstrates that the coding capabilities of text-only LLMs can effectively assist multimodal learning and unleash the potential of vision-language models for real-world applications.

\section*{Limitation}
The effectiveness of synthetic data depends heavily on the quality and diversity of the prompts and rendering pipelines used for data generation. 
For highly specialized or underrepresented domains, generating sufficiently diverse data remains challenging and may require careful prompt engineering or additional customization of rendering tools.
Targeted synthetic data generation may be essential for certain tasks to achieve adequate performance, and ensuring relevance and coverage still requires domain-specific expertise. Synthetic data also may not fully capture the complexity of real-world data in some scenarios. Therefore, improving the diversity and realism of synthetic data to better support models in highly variable or evolving domains is a reasonable avenue for future research.
Finally, our current synthetic data is limited to English and may require further extension for multilingual support.

\section*{Ethical Statement}

To the best of our knowledge, this work presents no significant ethical concerns. We note, however, that the use of synthetic data can propagate biases present in the generation model used. Conversely, synthetic data can also help mitigate biases and expand coverage, as demonstrated in this work, by greatly expanding the domains present in vision-language instruction-tuning training data to yield stronger generalized performance.

\section*{Acknowledgement}
This work was done during Yue Yang's internship at the PRIOR team of Ai2.
This research is supported in part by the Office of the Director of National Intelligence (ODNI), Intelligence Advanced Research Projects Activity (IARPA), via the HIATUS Program contract \#2022-22072200005, and the Defense Advanced Research Projects Agency's (DARPA) SciFy program (Agreement No. HR00112520300), and gifts from the UPenn ASSET center and Ai2. 
The views and conclusions contained herein are those of the authors and should not be interpreted as necessarily representing the official policies, either expressed or implied, of ODNI, IARPA, DARPA, or the U.S. Government. 
The U.S. Government is authorized to reproduce and distribute reprints for governmental purposes, notwithstanding any copyright annotation therein.

\bibliography{custom}

\clearpage
\appendix

\section{Implementation Details}
\subsection{Prompts}
We provide the prompt templates in Figure \ref{fig:prompt-html} for the \textit{HTMLDocumentPipeline} as an example to illustrate the prompts used across our code-guided synthetic data generation pipelines.

\begin{figure}[!ht]
    \centering
    \begin{center}
    
    \begin{tcolorbox} [top=3pt,bottom=3pt, left=3pt, right=3pt, width=\linewidth, boxrule=1pt]
    {\scriptsize {\fontfamily{phv}\selectfont    
    \underline{\textbf{Topic Generation:}}
    You are an expert in document generation and have a broad knowledge of different topics.
    
My persona is: ``\textbf{PERSONA}''
I want you to generate \textbf{NUM\_TOPICS} topics for \textbf{FIGURE\_TYPE} that I will be interested in or I may see during my daily life given my persona.

Here are the requirements:

1. Each topic is a high-level summary of the contents in \textbf{FIGURE\_TYPE} with some design details, e.g., "the utility bill for the month of January 2022 with a detailed breakdown of charges".

2. The topics should be diverse to help me generate varied documents. Each topic should be unique and not overlap with others.

3. The topics are conditioned on the document type. Please ensure the topics you provided can be best visualized in "\textbf{FIGURE\_TYPE}".

4. All topics must be in English, even if the persona is non-English.

5. List \textbf{NUM\_TOPICS} topics for "\textbf{PERSONA}" and separate them with a | character, e.g., topic1 | topic2 | ...... | topicN.

Do not include any additional text at the beginning or end of your response.}    \par}
    \end{tcolorbox}

\begin{tcolorbox} [top=3pt,bottom=3pt, left=3pt, right=3pt, width=\linewidth, boxrule=1pt]
    {\scriptsize {\fontfamily{phv}\selectfont   
    \underline{\textbf{Data Generation:}}
    You are an expert in content creation and have broad knowledge about various topics.
    
My persona is: "\textbf{PERSONA}"
I need some materials about "\textbf{TOPIC}", which can be used to generate a \textbf{FIGURE\_TYPE}. 

Here are the requirements:

1. The materials should be related to the topic and customized according to my persona. Its structure must be suitable for the \textbf{FIGURE\_TYPE}.

2. The materials should be realistic, and the contents should be named using real-world entities. Do not use placeholder names like xxA, xxB, etc. Do not use template data like [Name], [Date], etc.

3. The materials should be diverse and contain information from different aspects of the topic to ensure the document is informative.

4. Do not provide too many materials. Just provide key pieces of information that are essential for a **one-page document.**

5. All materials must be in English, even if the persona is non-English.

Please provide the materials in JSON format without additional text at the beginning or end.}    \par}
    \end{tcolorbox}

\begin{tcolorbox} [top=3pt,bottom=3pt, left=3pt, right=3pt, width=\linewidth, boxrule=1pt]
    {\scriptsize {\fontfamily{phv}\selectfont    
    \underline{\textbf{Code Generation:}}
    You are an expert web designer and are good at writing HTML to create documents.
    
My persona is: "\textbf{PERSONA}"
I have some materials about \textbf{TOPIC} which can be used to generate a \textbf{FIGURE\_TYPE}.

Here are the materials (JSON format):

<data>
\textbf{DATA}
</data>

Please use HTML and CSS to generate a \textbf{FIGURE\_TYPE} using the data provided. 

Here are the requirements:

1. **Style Requirements**: Feel free to use any CSS framework, libraries, JavaScript plugins, or other tools to create the document.

    (1) Try to be creative and make the web page style, fonts, colors, borders and visual layout unique with CSS. Taking persona, topic, and document type into consideration when designing the document.
    
    (2) Select the appropriate design scale (e.g., margins, page size, layout, etc) to ensure the information in the document is clear and easy to understand, with no text overlapping, etc.
    
    (3) **Do not make the page too long or too sparse.** All contents should be in **one page**. This is very important.
    
2. **Code Requirements**: 

    (1) You need to hardcode the provided data into the HTML script to generate the document. Be careful with the syntax and formatting of the HTML.
    
    (2) Put everything in one HTML file. Do not use external CSS or JavaScript files.

3. **Output Requirements**:
    Put ```html at the beginning and ``` at the end of the script to separate the code from the text.

Please don't answer with any additional text in the script, your whole response should be the HTML code which can be directly executed.}    \par}
    \end{tcolorbox}

    \end{center}
\end{figure}

\begin{figure}[!h]
    \centering
    \begin{center}

\begin{tcolorbox} [top=3pt,bottom=3pt, left=3pt, right=3pt, width=\linewidth, boxrule=1pt]
    {\scriptsize {\fontfamily{phv}\selectfont    
    \underline{\textbf{Instruction Generation:}}
    You are an expert in data analysis and good at asking questions about documents.
My persona is: "{persona}"
I want you to generate some question-answer pairs of a \textbf{FIGURE\_TYPE} about \textbf{TOPIC}, which I would ask.
Instead of showing the document, I provide the data and the code that generates the document.

<data>
\textbf{DATA}
</data>
<code>
\textbf{CODE}
</code>

Please come up with a list of *reasonable questions* that people will ask when they see the rendered document. Here are the requirements:

1. **Question Types**: All questions should be short-answer questions that are answerable based on the visual information in the document. All questions can be answered with a single word, phrase, or number. (as short as possible)

    (1) **Information Retrieval questions** ask for specific information in the document, such as numbers, names, dates, titles, etc. The questions should cover different aspects (areas) of the document. This is the most common type of question.
    
    (2) **Reasoning questions** require reasoning over multiple information in the document. These questions should be more challenging and require a deeper understanding of the document.
    
    (3) **Document Type-specific questions** are questions that are specific and unique to this document type \textbf{FIGURE\_TYPE}. These questions should be tailored to the content and structure of the document.

2. **Response Format**: Use | to separate the question, explanation, and concise answer for each example. 

    (1) Follow this format: question | explanation | concise answer, e.g., what is the total revenue? | The total revenue is the sum of all revenue sources in the document, which is \$2000 + \$3000 + \$5000 = \$10000. | \$10000 
    
    (2) Separate the question-answer pairs by double newlines. 
    question1 | explanation1 | answer1 
    
    question2 | explanation2 | answer2...
    
    (3) Do not provide too many questions, 5-10 questions are enough. Focus on the diversity and quality of the questions. Try to cover different aspects of the document.
    
    (4) The concise answer should be as short as possible and directly answer the question. The answer should be faithful and exactly the same as what you would expect to see in the document, don't rephrase it. All words in the answer should be processed in natural language, no coding terms/characters.
    
Please follow the format strictly and do not include any additional text at the beginning or end of your response.}    \par}
    \end{tcolorbox}

    \end{center}
    \vspace{-0.3cm}
    \caption{Prompt templates used for HTML Document Pipeline, including all four stages of generation: topic, data, code, and instruction.}
    \label{fig:prompt-html}
    \vspace{-0.3cm}
\end{figure}

\subsection{Rendering Tools and Pipelines}
We design 20 generation pipelines built on 11 rendering tools to support the creation of nine categories of text-rich images:(1) \textbf{Charts}: Matplotlib VegaLite, Plotly, LaTeX, HTML;
(2) \textbf{Documents}: LaTeX, HTML;
(3) \textbf{Tables}: LaTeX, Matplotlib, Plotly, HTML;
(4) \textbf{Diagrams}: Graphviz, LaTeX, Mermaid;
(5) \textbf{Math Problems}: LaTeX;
(6) \textbf{Vector Graphics}: SVG, Asymptote;
(7) \textbf{Music Sheets}: LilyPond;
(8) \textbf{Electrical Circuits}: LaTeX;
(9) \textbf{Chemical Structures}: Rdkit. 
In addition, we implement a separate pipeline for generating pointing data using HTML as the rendering tool.

\subsection{Queries to Construct CoSyn-400K} \label{appendix: query}
Since CoSyn accepts textual queries to control data generation, we use a diverse set of queries for each type of text-rich image to ensure broad domain coverage. 
Below are some examples of the queries used to generate CoSyn-400K:
\begin{itemize}[noitemsep, topsep=0pt, leftmargin=*]
    \item \textbf{Charts}: Bar, Line, Pie, Diverge bar, Bubble, Scatter, Histogram, Area, Box plot, Heatmap, Error bar, Radar chart, Rose chart, Stem plot, Stairs plot, Violin chart, 2D contour, Distplots, Log plot, Ternary plots/contour, Candlestick charts, Time series, etc. (\textbf{51} queries in total)
    
    \item \textbf{Documents}: Letter, Form, Report, Receipt, Invoice, Restaurant menu, Newsletter, Schedule, Manual, Brochure, Transaction document, Agenda, Memo, Financial report, Telephone records, Note, Budget, Meeting minutes, Bill, Catalog, Email, Fax, Policy document, Resume, Infographics, Process infographic, Statistical infographic, etc. (\textbf{107} queries in total)
    
    \item \textbf{Math Problems}: Algebra, Counting, Probability, Geometry, Number theory, Precalculus, Prealgebra, Intermediate Algebra, Statistics, Functions, Complex numbers, Logarithms, Inequalities, Linear equations, Exponents, Series, College Algebra, Calculus, Advanced calculus, Linear algebra, Solid geometry, Analytic geometry, Polynomial arithmetic, etc. (\textbf{110} queries in total)

    \item \textbf{Tables}: Financial table, Simple table, Pivot table, Comparison table, Timeline table, Decision table, Truth table, Lookup table, Periodic table, Statistical table, Timetable, Hierarchical table, Matrix table, Contingency table, Logarithmic table, Correlation table, etc. (\textbf{35} queries in total)

    \item \textbf{Diagrams}: Flow chart, Directed graph, Undirected graph, Decision tree, Mind map, Gantt charts, Finite state machine, Quadrant chart, Chord diagrams, Network diagrams, Sankey diagram, Entity relationship diagram, Sequence diagrams, Bottom-up flow chart, Timeline, State diagram, Concept map, Family tree, Programming flowchart, etc. (\textbf{34} queries in total)

    \item \textbf{Vector Graphics}: Visual intelligence test, Spatial intelligence test, Geometry, Solid geometry, Analytic geometry, Polynomial graphs, Trigonometry, Polar coordinates, Coordinate system, Topology, Graph theory, Plane geometry, Functions, Calculus, Vectors, Angles, Perimeter and area problems, etc. (\textbf{36} queries in total)

    \item \textbf{Sheet Music}: Classical, Pop, Rock, Jazz, Blues, Hip Hop, Rap, Electronic, Country, Folk, Rhythm and blues, Soul, Reggae, Metal, Punk, Theme, Dance, etc. (\textbf{34} queries in total)
    
    \item \textbf{Electrical Circuits}: Series, Parallel, Hybrid, Household appliances, Industrial appliances, Mobile device, Low-power appliances, High-power appliances, etc. (\textbf{30} queries in total)

    \item \textbf{Chemical Structures}: Drug, Organic, Inorganic, Protein, Acids, Bases, Gases, Liquids, Solids, Oxidizers, Flammable liquids, Toxic chemicals, Hazardous chemicals, Aromatic compounds, Aliphatic compounds, Polymers, Metals, Alloys, Electrolytes, etc. (\textbf{100} queries in total)

\end{itemize}

\subsection{Academic Datasets}
During the supervised fine-tuning stage, we include academic datasets in addition to our synthetic datasets. Below, we provide details on the size of these datasets and the evaluation metrics used.

\smallbreak
\noindent \textbf{Dataset Size.}
The number in parentheses indicates the number of training images for each dataset: ChartQA (28.3K), DocVQA (39.5K), InfographicVQA (23.9K), AI2 Diagrams (11.4K), TextVQA (34.6K), VQAv2 (82.8K), GQA (72.1K), OK-VQA (9.0K), OCR-VQA (166.0K), A-OKVQA (17.1K), ScienceQA (6.2K), TabMWP (23.1K), ST-VQA (18.9K), TallyQA (133.0K), DVQA (200.0K), FigureQA (100.0K), PlotQA (160.0K).
We downsample some very large synthetic datasets, such as DVQA, FigureQA, and PlotQA, to balance the dataset size. 
In total, we use approximately 1.1M images from academic datasets.

\smallbreak
\noindent \textbf{Evaluation Metrics.} We adopt their official evaluation metrics for the seven text-rich datasets. 
For ChartQA, we use relaxed correctness, which allows a 5\% difference for float number answers. 
For DocQA and InfoQA, we report Average Normalized Levenshtein Similarity (ANLS).
For TableVQA, we report the average performance across the four subsets (VTabFact, VWTQ, VWTQ-Syn, FinTabNetQA) using the metrics provided in this \href{https://github.com/naver-ai/tablevqabench}{repo}.
We report the multiple choice accuracy for AI2D, VQA score \cite{balanced_vqa_v2} for TextVQA, and SQuAD F1 score \cite{rajpurkar-etal-2018-know} for ScreenQA.

\begin{table*}[!ht]
    \small
    \centering
    \begin{tabular}{lcccccccccc}
        \toprule
        \textbf{Prompt Type} & \textbf{ChartQA} & \textbf{DocVQA} & \textbf{InfoVQA} & \textbf{TableVQA} & \textbf{AI2D} & \textbf{TextVQA} & \textbf{ScreenQA} & \textbf{NutritionQA}\\
        \midrule
        CoT & \textbf{86.3} & 87.4 & 63.8 & \textbf{65.8} & 86.0 & 70.9 & 79.0 & \textbf{76.0}\\
        Short Answer & 83.1 & \textbf{90.0} & \textbf{70.5} & 64.3 & \textbf{91.9} & \textbf{82.0} & \textbf{80.1} & 62.0\\
        \bottomrule
    \end{tabular}
    \vspace{-.1cm}
    \caption{\textbf{Alation of using chain-of-thought (CoT) in prompts.} CoT means letting the model provide reasoning steps before giving the final answer. Short Answer prompts the model to answer with as few words as possible.}
    \label{tab:cot}
\end{table*}
\begin{table*}[!t]
    \small
    \centering
    \begin{tabular}{lcccccccccc}
        \toprule
        \textbf{FT Data} & \textbf{ChartQA} & \textbf{DocVQA} & \textbf{InfoVQA} & \textbf{TableVQA}$^\dagger$ & \textbf{AI2D} & \textbf{TextVQA} & \textbf{ScreenQA}$^\dagger$ & \textbf{Average} \\
        \midrule
        Aux only$^*$ & 60.7 & 56.2 & 39.7 & 43.1 & 81.7 & 68.5 & 61.3 & 58.7\\
        Syn only$^*$ & 79.4 & 80.5 & 60.1 & 64.4 & 68.6 & 63.6 & 76.6 & 70.5\\
        Aux + Syn$^*$ & 80.8 & 82.9 & 59.8 & 64.9 & 83.9 & 72.7 & 78.1 & 74.7\\ \midrule
        Eval only & 77.4 & 87.4 & 63.8 & 51.8 & 91.3 & 81.1 & 78.1 & 75.9 \\
        Eval + Aux & 81.4 & 87.9 & 68.2 & 53.6 & 91.6 & 81.8 & 77.0 & 77.3\\
        Eval + Aux + Syn & \textbf{86.3} & \textbf{90.0} & \textbf{70.5} & \textbf{65.8} & \textbf{91.9} & \textbf{82.0} & \textbf{80.1} & \textbf{80.9}\\
        \bottomrule
    \end{tabular}
    \vspace{-.1cm}
    \caption{\textbf{Alation of the data selection for supervised fine-tuning.} Aux, Syn, and Eval stand for auxiliary, synthetic, and evaluation datasets, respectively. The rows with $^*$ represent zero-shot models (without using any training examples from any of the evaluation datasets). The datasets with $^\dagger$ are test-only datasets (no training splits), which means all numbers on these datasets are zero-shot performance.}
    \label{tab:train_data_ablation}
    \vspace{-.3cm}
\end{table*}

\subsection{Training Details}

\smallbreak
\noindent \textbf{Image Preprocessing.} We adopt the same image preprocessing as Molmo \citep{deitke2024molmo}, where each input image is cropped into multiple overlapping crops before being encoded by CLIP. 
During training, we limit the maximum number of crops to 12, but we increase it to 25 at testing time to accommodate the high resolution of text-rich images. 
This strategy boosts the inference performance without increasing training costs.

\smallbreak
\noindent \textbf{Hyper Parameters.} We set the maximum sequence length for training to 2304 tokens. 
We use the same learning rate of 1e-6 for the MLP connector, LLM, and visual encoder, with batch size 32. 
The best-performing model is trained for 60K steps with 200 warm-up steps and a cosine scheduler with an end factor of 0.1. 
All experiments are run on a single TPU v3-128.

\section{Additional Analysis}
We conduct additional analyses below to investigate further why our synthetic data can effectively enhance vision-language models.

\smallbreak
\noindent \textbf{Quantify the contributions of synthetic data.} Table \ref{tab:train_data_ablation} presents the performance across benchmarks using different combinations of supervised fine-tuning data. 
A clear trend shows that synthetic data significantly contributes in both zero-shot and supervised settings. 
Adding our synthetic data consistently boosts performance on each benchmark.

\smallbreak
\noindent \textbf{The impact of Chain-of-thought reasoning.} We compare the performance of CoT and short-answer prompts in Table \ref{tab:cot}.
CoT reasoning improves performance on ChartQA, TableVQA, and NutritionQA, where questions require multi-hop and mathematical reasoning that aligns with the findings in language tasks \cite{sprague2024cot}.
However, short-answer prompts yield better results for the other five datasets due to their annotation biases favoring concise responses. 
CoT responses tend to be more verbose, which may not match the ground-truth answers exactly, resulting in a performance drop.

\smallbreak
\noindent \textbf{Document Pointing Task.} To further validate the effectiveness of our synthetic pointing data, we introduce DocPointQA\footnote{\href{https://huggingface.co/datasets/yyupenn/DocPointQA}{https://huggingface.co/datasets/yyupenn/DocPointQA}}, a new pointing task with 300 question-point pairs annotated from the DocVQA validation set (Figure \ref{fig: docpointqa}).
We compare models trained on human-annotated PixMo-point data (155K examples), our synthetic pointing data (65K examples), and their combination. 
Since DocPointQA requires multiple-point answers, we report precision, recall, F1 score, and L2 distance (lower is better) after mapping predicted points to ground truth, following the same setup as Molmo \cite{deitke2024molmo}.
As shown in Table \ref{tab:doc_point}, the model trained on our synthetic data outperforms the one trained on PixMo-point. 
Performance improves even further when both datasets are combined, demonstrating the effectiveness of synthetic data in enhancing the pointing capabilities of vision-language models.

\begin{table}[H]
    \small
    \centering
    \setlength{\tabcolsep}{4pt}
    \begin{tabular}{lcccc}
        \toprule
        \textbf{Pointing Data} & \textbf{Precision} & \textbf{Recall} & \textbf{F1} &   \textbf{Distance $\downarrow$} \\
        \midrule
        PixMo-point & 49.7 & 49.3 & 52.7 & 17.3 \\ \midrule
        Synthetic (Ours) & 63.8 & 66.1 & 62.8 & 9.2 \\
        Combined (Ours) & \textbf{69.9} & \textbf{70.6} & \textbf{70.7} & \textbf{8.8} \\
        \bottomrule
    \end{tabular}
    \vspace{-.1cm}
    \caption{\textbf{Zero-shot Pointing on DocPointQA.} We compare the models trained on different pointing data. Combined stands for combining PixMo-point (human-annotated) \cite{deitke2024molmo} with our synthetic data.}
    \label{tab:doc_point}
\end{table}

\section{Qualitative Examples} \label{appendix: example}
\begin{figure*}[!ht]
    \centering
    \includegraphics[width=\textwidth]{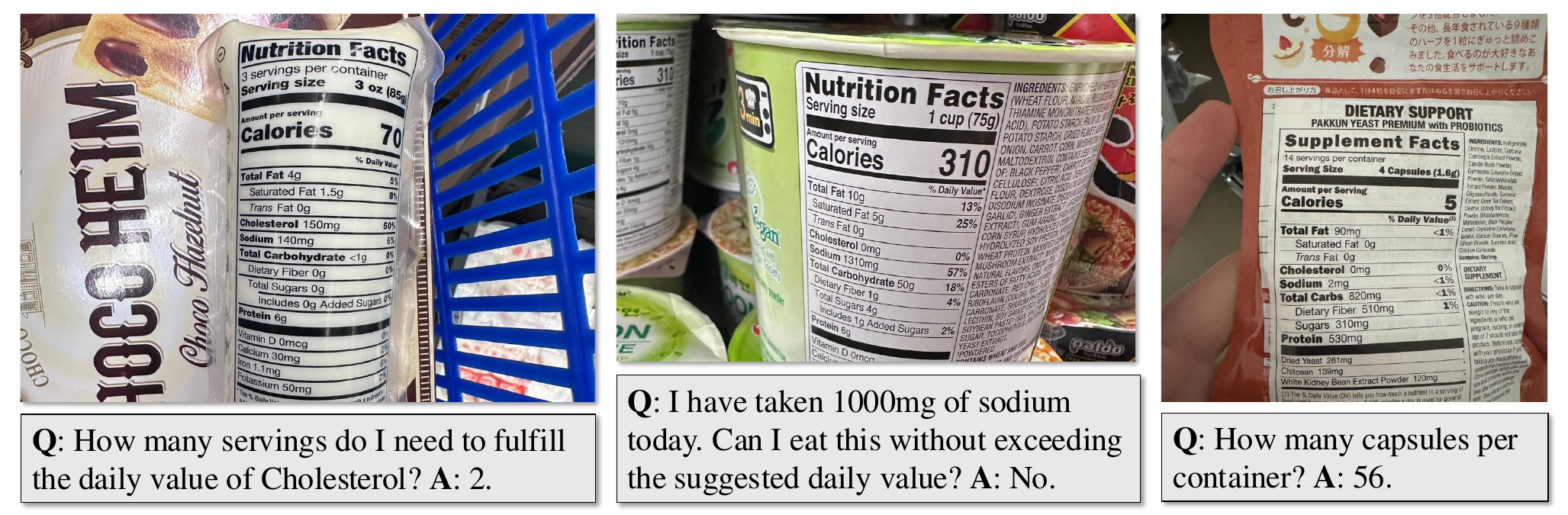}
    \caption{Examples from our newly collected \textbf{NutritionQA} dataset.}
    \label{fig: nurition_qa}
    \vspace{-0.3cm}
\end{figure*}

\begin{figure*}[!ht]
    \centering
    \includegraphics[width=\textwidth]{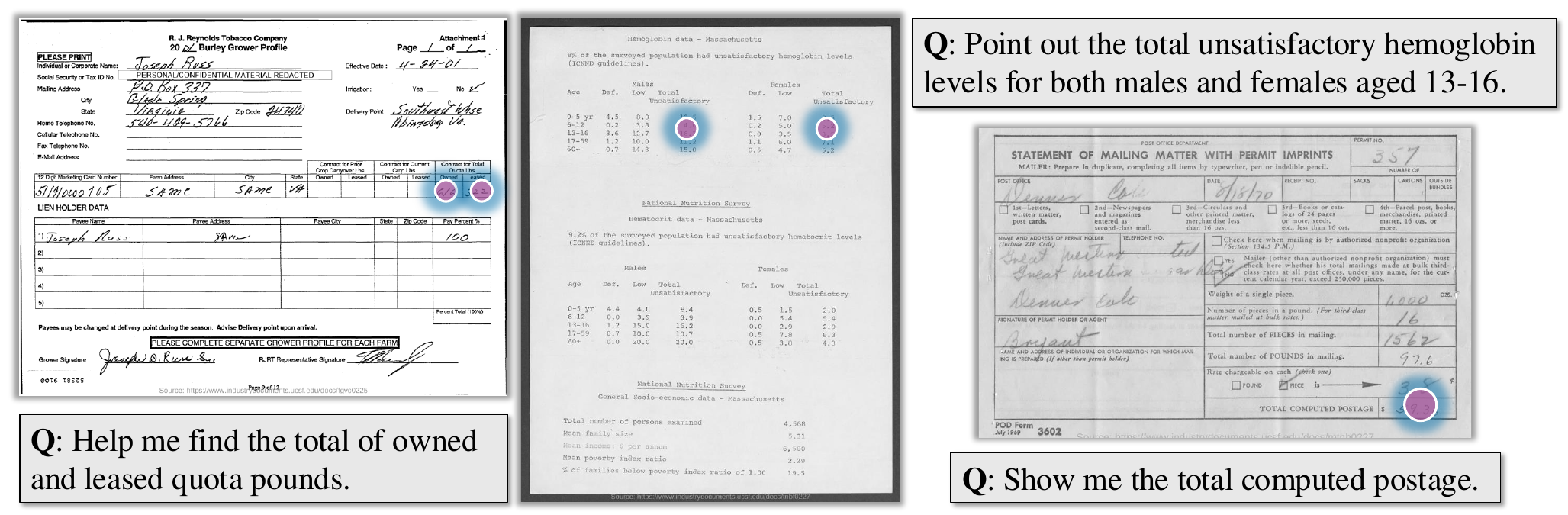}
    \caption{Examples from our newly collected \textbf{DocPointQA} dataset.}
    \label{fig: docpointqa}
    \vspace{-0.3cm}
\end{figure*}

\begin{figure*}[!ht]
    \centering
    \includegraphics[width=\textwidth]{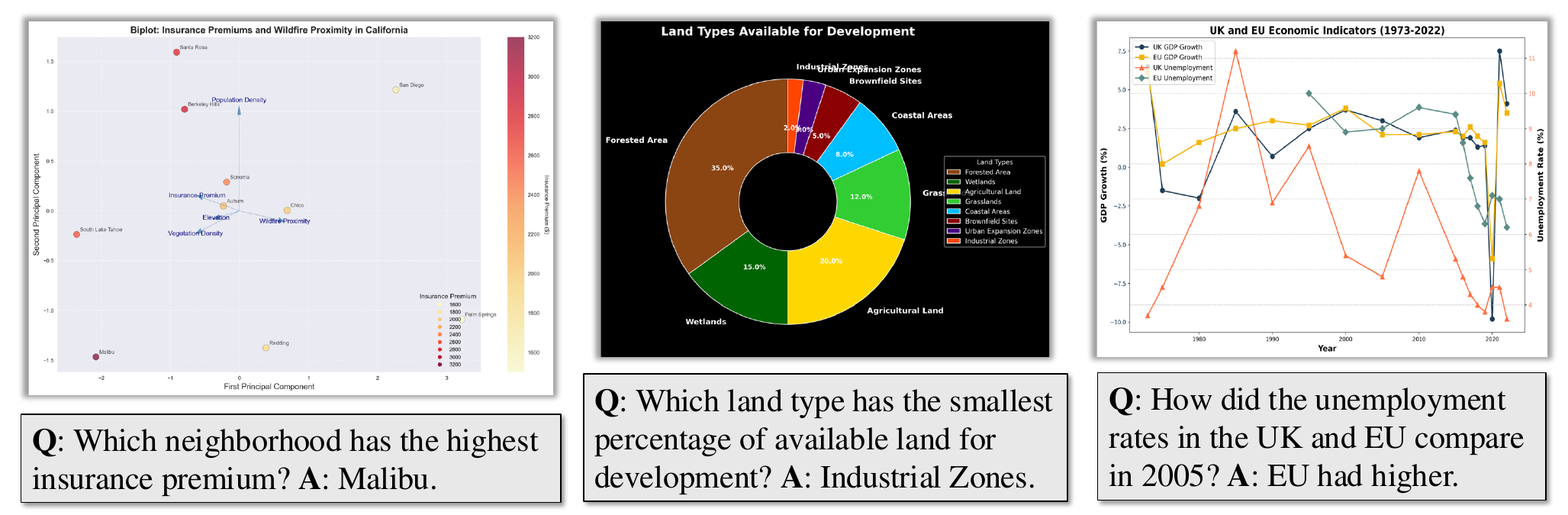}
    \caption{Randomly selected examples from our synthetic \textbf{chart} data.}
    \label{fig: chart_example}
    \vspace{-0.3cm}
\end{figure*}

\begin{figure*}[!t]
    \centering
    \includegraphics[width=\textwidth]{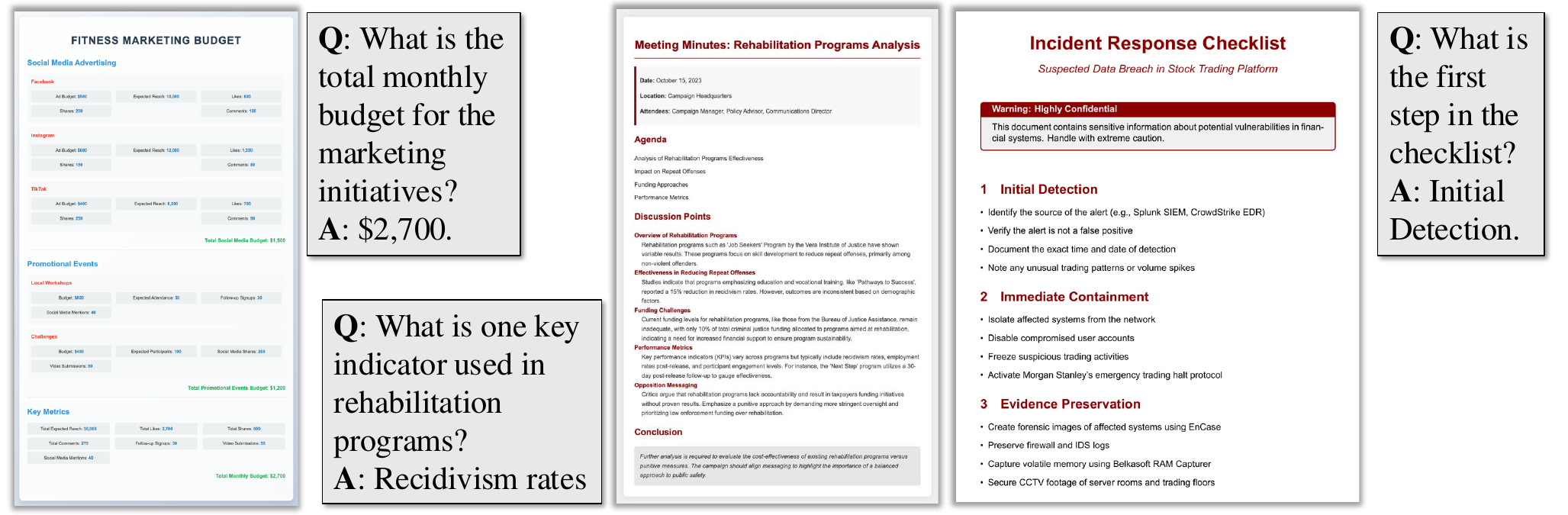}
    \caption{Randomly selected examples from our synthetic \textbf{document} data.}
    \label{fig: document_example}
    \vspace{-0.3cm}
\end{figure*}

\begin{figure*}[!t]
    \centering
    \includegraphics[width=\textwidth]{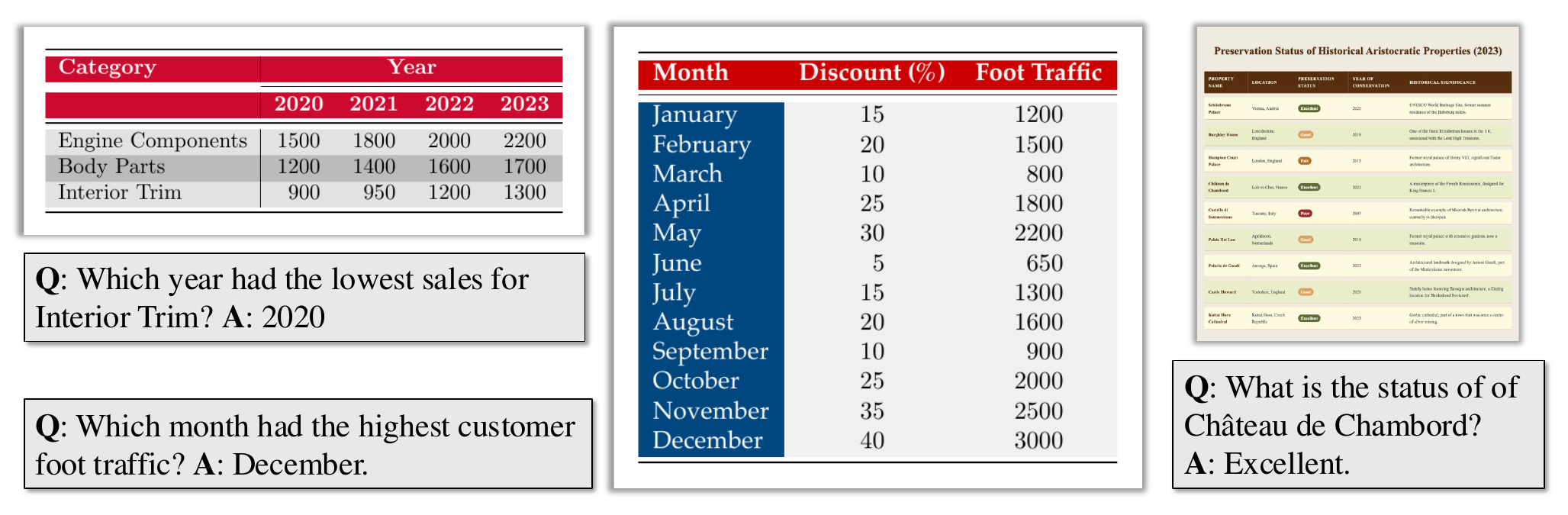}
    \caption{Randomly selected examples from our synthetic \textbf{table} data.}
    \label{fig: table_example}
    \vspace{-0.3cm}
\end{figure*}

\begin{figure*}[!t]
    \centering
    \includegraphics[width=\textwidth]{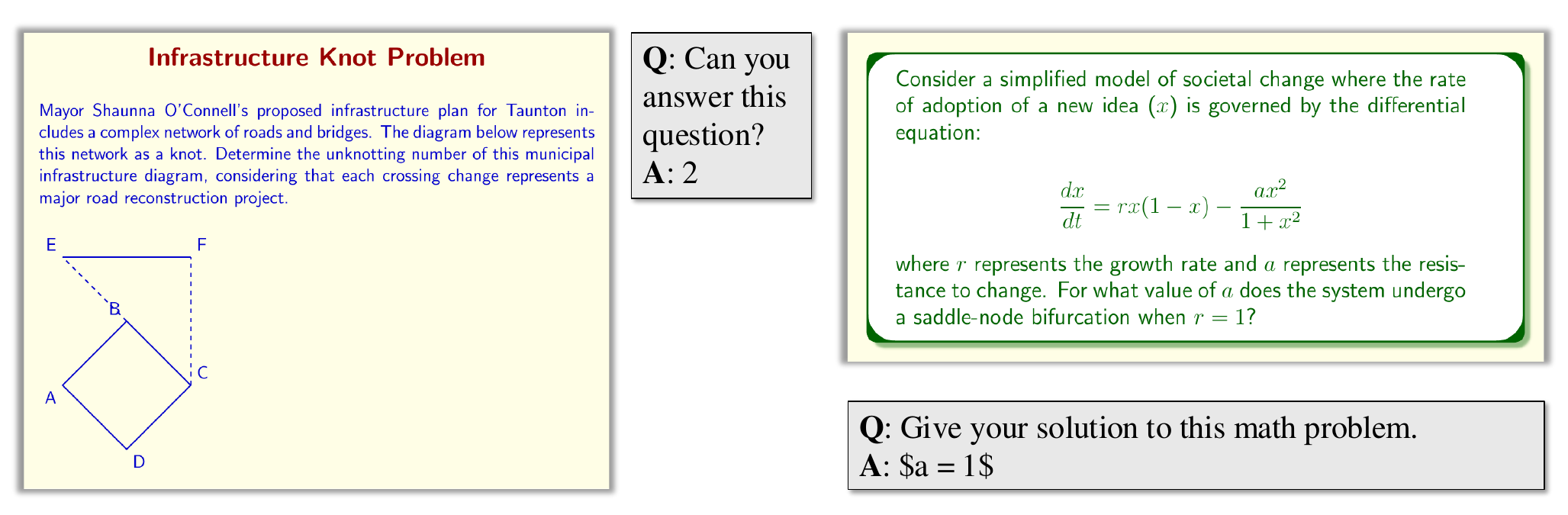}
    \caption{Randomly selected examples from our synthetic \textbf{math} data.}
    \label{fig: math_example}
    \vspace{-0.3cm}
\end{figure*}

\begin{figure*}[!t]
    \centering
    \includegraphics[width=\textwidth]{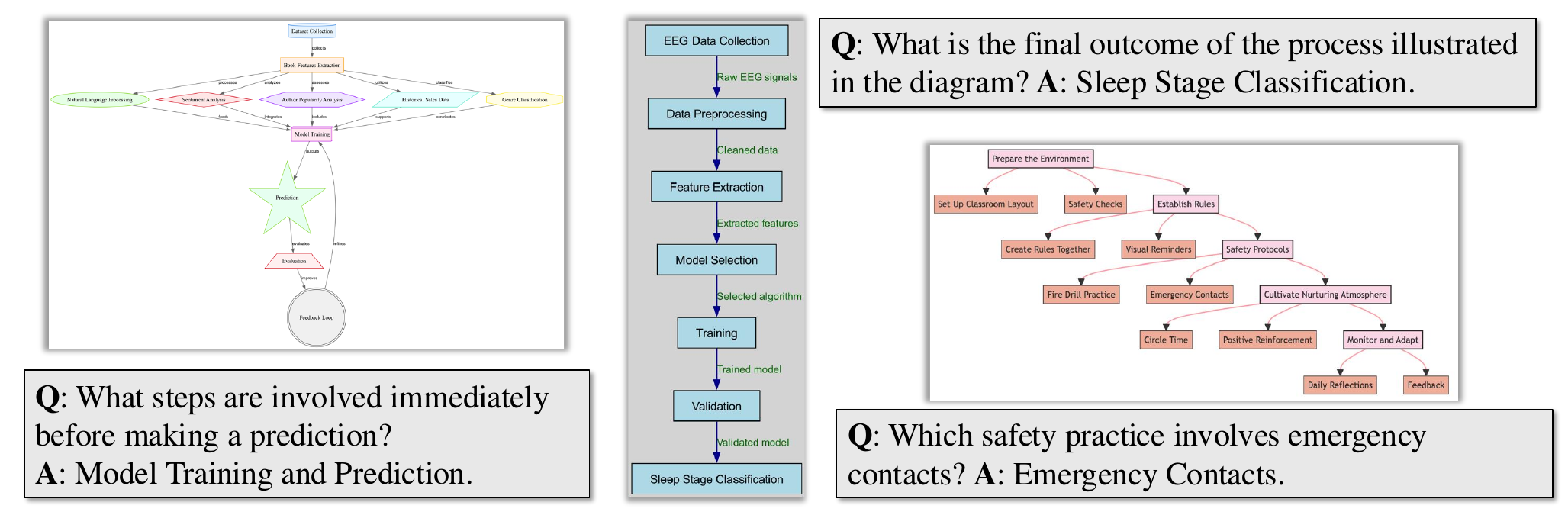}
    \caption{Randomly selected examples from our synthetic \textbf{diagram} data.}
    \label{fig: diagram_example}
    \vspace{-0.3cm}
\end{figure*}

\begin{figure*}[!t]
    \centering
    \includegraphics[width=\textwidth]{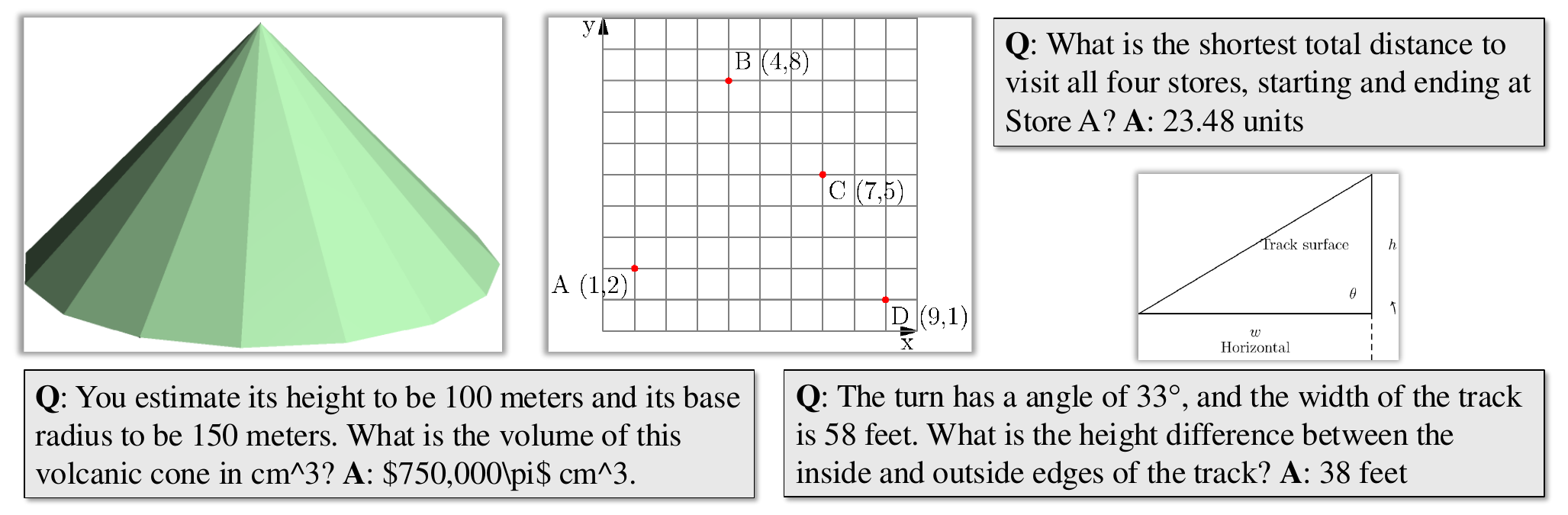}
    \caption{Randomly selected examples from our synthetic \textbf{vector graphic} data.}
    \label{fig: graphic_example}
    \vspace{-0.3cm}
\end{figure*}

\begin{figure*}[!t]
    \centering
    \includegraphics[width=\textwidth]{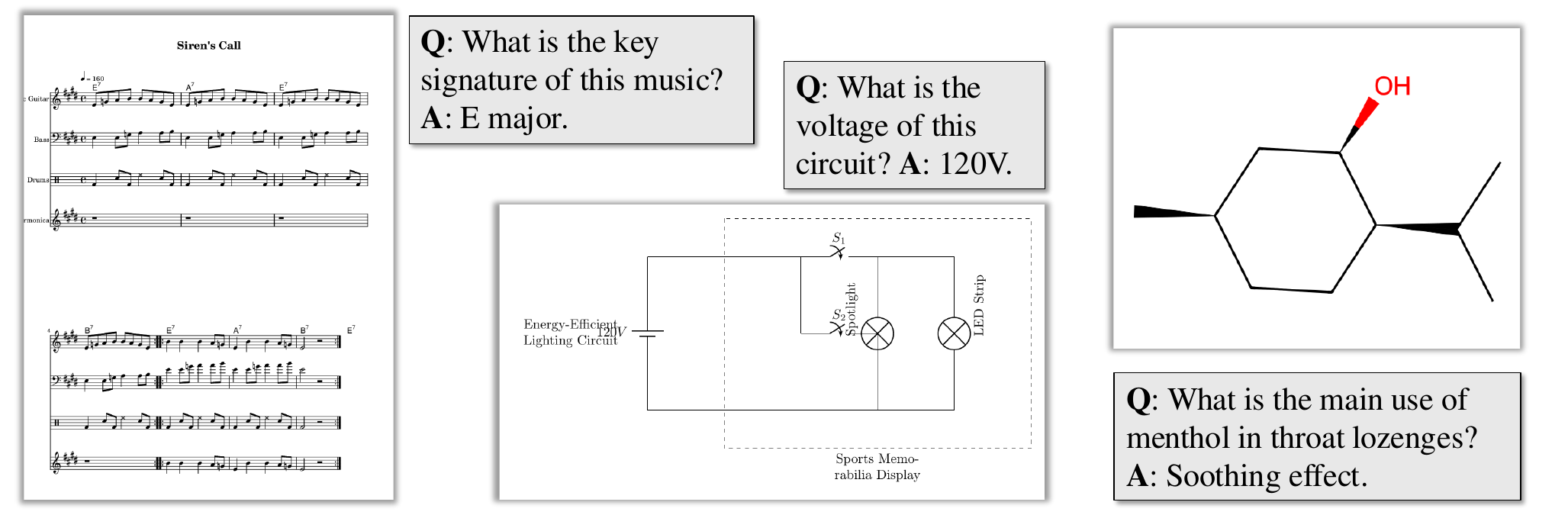}
    \caption{Randomly selected examples from our synthetic \textbf{sheet music}, \textbf{circuits} and \textbf{chemical structures}.}
    \label{fig: special_example}
    \vspace{-0.3cm}
\end{figure*}

\begin{figure*}[!t]
    \centering
    \includegraphics[width=\textwidth]{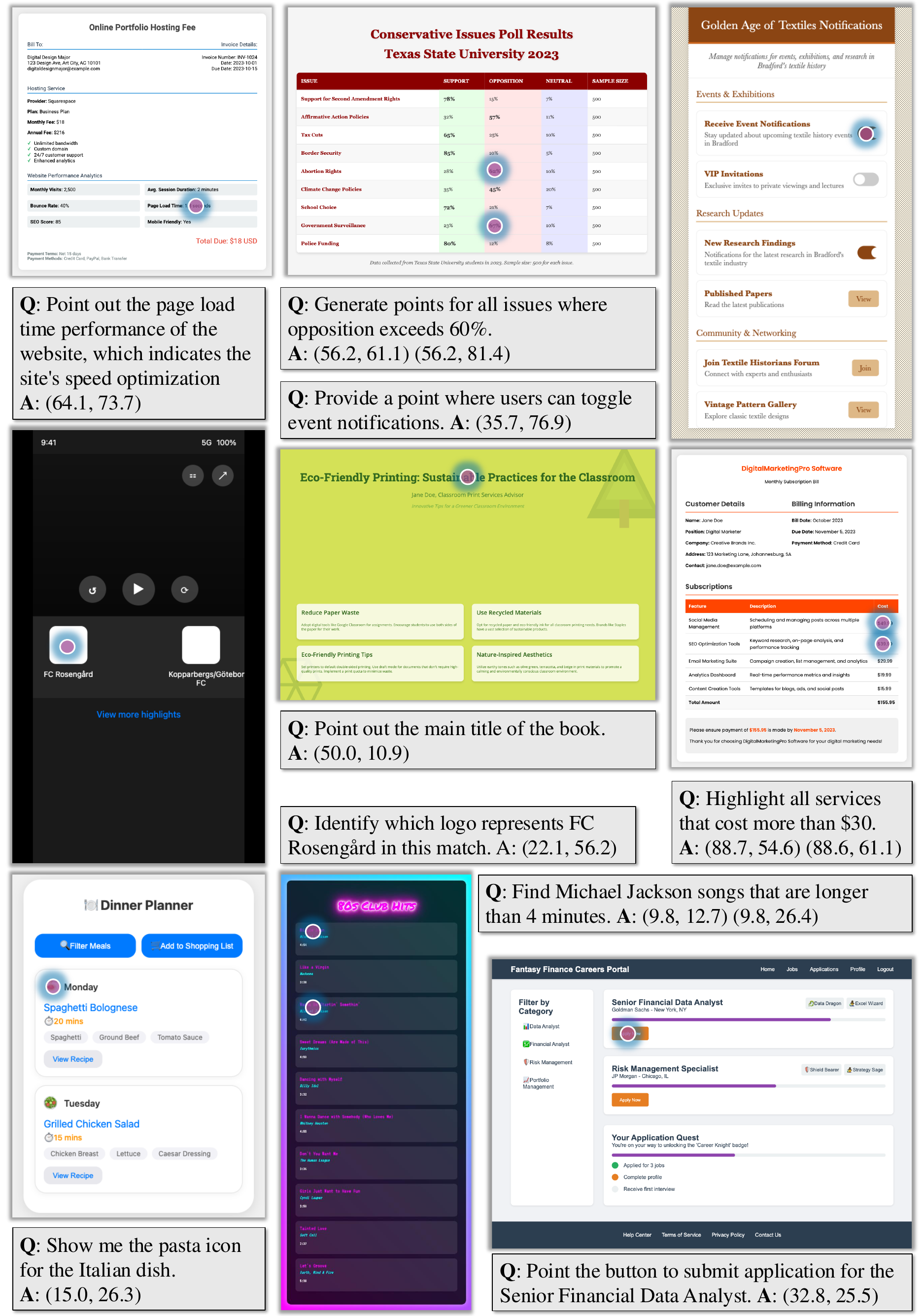}
    \caption{Randomly selected examples from our synthetic \textbf{pointing} data.}
    \label{fig: point_example}
    \vspace{-0.3cm}
\end{figure*}

Figure \ref{fig: nurition_qa} and \ref{fig: docpointqa} show the examples from our annotated NutritionQA\footnote{\href{https://huggingface.co/datasets/yyupenn/NutritionQA}{https://huggingface.co/datasets/yyupenn/NutritionQA}} and DocPointQA. 
Figures \ref{fig: chart_example} - \ref{fig: special_example} list examples from the 9 categories of synthetic text-rich images. Figure \ref{fig: point_example} illustrates examples from the synthetic pointing dataset.

\clearpage

\clearpage

\clearpage

\end{document}